\documentclass[10pt,twocolumn,letterpaper]{article}
\usepackage[accsupp]{axessibility}
\usepackage{iccv}
\usepackage{times}
\usepackage{epsfig}
\usepackage{graphicx}
\usepackage{amsmath}
\usepackage{amssymb}
\usepackage{textcomp}
\usepackage{enumitem}
\usepackage{caption}
\usepackage[skip=3pt]{subcaption}

\usepackage[breaklinks=true,bookmarks=false]{hyperref}

\iccvfinalcopy % *** Uncomment this line for the final submission

\def\iccvPaperID{2462} % *** Enter the ICCV Paper ID here

\ificcvfinal\fi

\newenvironment{packed_lefty_item}{
\begin{itemize}[leftmargin=*]
\vspace{-3pt}
  \setlength{\itemsep}{0pt}
  \setlength{\parskip}{0pt}
  \setlength{\parsep}{0pt}
  \setlength{\topsep}{-5pt}
  \setlength{\partopsep}{0pt}
}{\end{itemize}\vspace{-6pt}}

\begin{document}

%%%%%%%%% TITLE
\title{GTT-Net: Learned Generalized Trajectory Triangulation}
\author{Xiangyu Xu \hspace{0.65cm} Enrique Dunn\\
Stevens Institute of Technology\\
{\tt\small \{xxu24, edunn\}@stevens.edu}
}

\maketitle
\ificcvfinal\thispagestyle{empty}\fi

%%%%%%%%% ABSTRACT
\begin{abstract}
   We present GTT-Net, a supervised learning framework for the reconstruction of sparse dynamic 3D geometry. We build on a graph-theoretic formulation of the generalized trajectory triangulation problem, where non-concurrent multi-view imaging geometry is known but global image sequencing is not provided. GTT-Net learns pairwise affinities modeling the spatio-temporal relationships among our input observations and leverages them to determine 3D geometry estimates. 
   Experiments reconstructing 3D motion-capture sequences show GTT-Net outperforms the state of the art in terms of accuracy and robustness. Within the context of articulated motion reconstruction, our proposed architecture is 1) able to learn and enforce semantic 3D motion priors for shared training and test domains, while being 2) able to generalize its performance across different training and test domains. Moreover, GTT-Net provides a  computationally streamlined framework for trajectory triangulation with applications to multi-instance reconstruction  and event segmentation.
\end{abstract}

%%%%%%%%% BODY TEXT
\section{Introduction}
Trajectory triangulation aims to estimate multi-view sparse dynamic 3D geometry in the absence of concurrent observations. 
Recent advances in modeling and estimating the  spatio-temporal relationships among 2D observations have yielded solutions with increasing generality and effectiveness. 
However, such research efforts have focused on developing and exploiting geometric insights and formulations, relegating the analysis of higher-order semantic relationships among the geometric entities being estimated. This work addresses the data-driven explicit characterization and modeling of these properties within the context of generalized trajectory triangulation.

%Of particular interest in the design of abstract representations 
Learning to encode generic spatio-temporal relationships hinges on the  geometric reference being used and the scope of the analysis. The choice of geometric reference typically poses a dichotomy between Eulerian (e.g. field approach) vs. Lagrangian (e.g. particle approach) representations, where the former defines interactions among rigidly structured adjacency-based neighborhoods    (e.g. voxel laticces), the latter defines interactions based on generic notions of proximity (e.g. nearest-neighbor graphs). Although  scope  is tightly coupled to these interaction mechanisms, the efficiency vs. comprehensiveness  trade-offs between local and global analysis,  determine the efficacy of the learned models and representations. We target a  discrete-continuous local-global middle ground by 1) learning to approximate  pairwise affinities over all estimated geometric elements, through 2)  the use of sparse continuous convolutions.

\begin{figure}[t!] 
\centering
\begin{subfigure}[h!] {0.45\textwidth} % width of left 
    \centering
	\includegraphics[scale = 0.23]{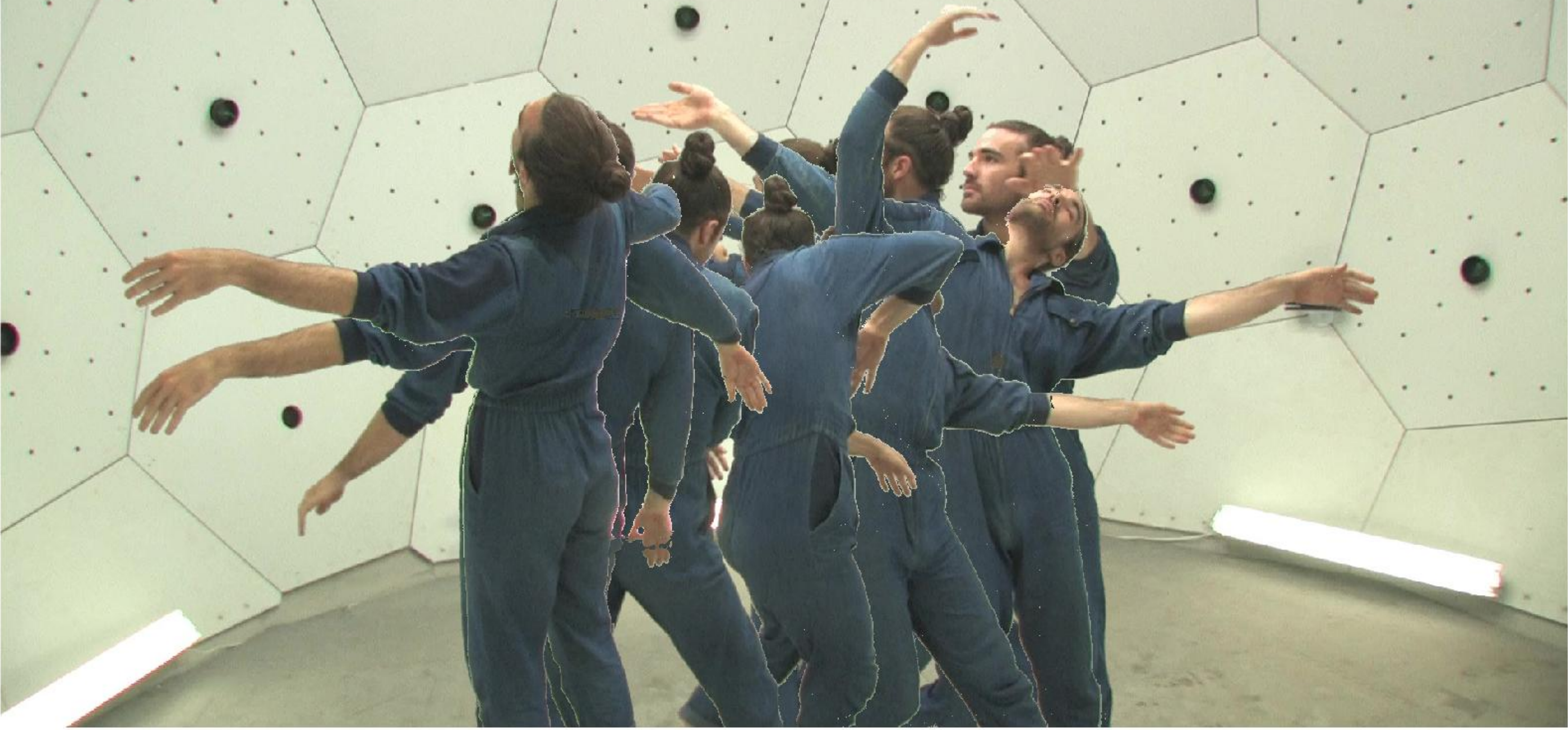}
\end{subfigure}
%\hspace{1em}
\begin{subfigure}[h!] {0.45\textwidth} % width of right
    \centering
	\includegraphics[scale = 0.32]{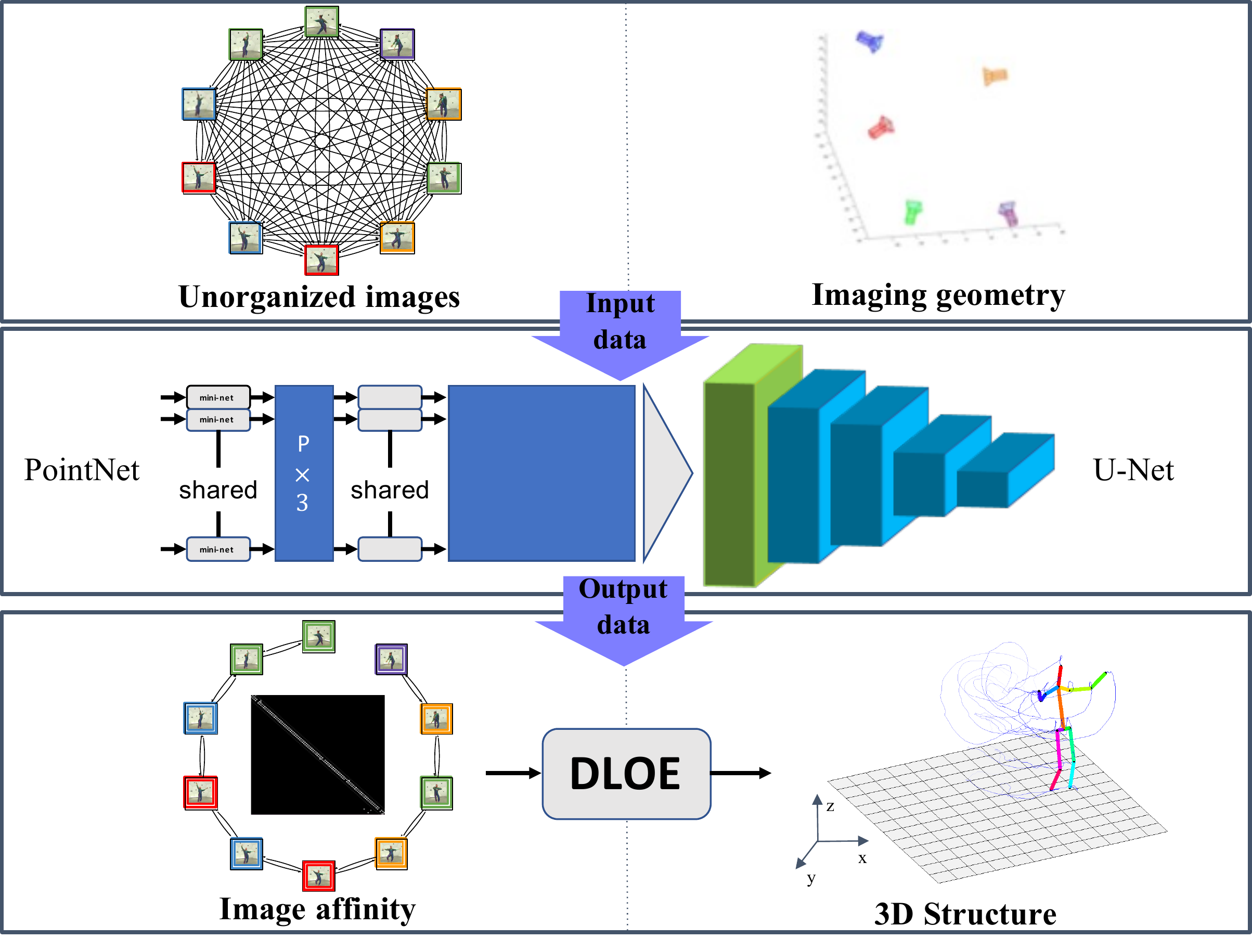}
\end{subfigure}

	\caption{GTT-Net Workflow. Input camera poses and  2D features are mapped to a latent space encoding a pairwise affinity matrix leveraged to estimate 3D geometry. }
	\label{fig:overview}
\end{figure}

%We integrate our approach into a 
Along these lines, the recent framework for generalized trajectory triangulation (GTT) described in \cite{Xu_2019_ICCV}, poses the estimation of such relationships in terms of the iterative continuous optimization of a graph-theoretic representation. However, said optimization offers relatively slow convergence and provides no straightforward mechanisms for codifying internal shape constraints or sequence-level motion priors. This work focuses on learning to synthesize a global shape affinity matrix directly from input 3D geometry to integrate with and  leveraging the representation and formulation used in \cite{Xu_2019_ICCV}, see Fig. \ref{fig:overview}. 
Our contributions are:
\begin{packed_lefty_item}
    \item{} A learning-based solution to the joint reconstruction and sequencing problems from multi-view image capture.
    \item{} A generalizable learning and representation framework applicable across diverse input shape domains.
%    \item{} Incorporating semantic 3D motion priors into the geometric estimation process through supervision.
    \item{} An efficient and flexible cascaded training framework applicable across diverse types of supervisory information.
\end{packed_lefty_item}

%This paper is organized as follows: We first review ralted work on trajectory triangulation and the recent relevant work on supervised learning for sparse 3D reconstruction. Next, we describe our generalized trajectory triangulation framework. Then we describe the details of our proposed network architecture and learning scheme. Finally, we present experimental results and conlusions.

%------------------------------------------------------------------------
\section{Related work}
%-------------------------------------------------------------------------
\subsection{Trajectory Triangulation} 
 Trajectory triangulation operates on the premise of  known cameras. However, the lack of concurrency requires enforcing estimation constraints to discriminate among the space of solutions compliant with the input observations.
 
\noindent {\bf Motion priors}. Avidan and Shashua \cite{avidan2000trajectory} enforced analytical linear and conical motion constraints upon the estimated 3D point trajectories from monocular capture. Extensions to these motion priors, include \cite{avidan1999trajectory,avidan2000trajectory,han2004reconstruction,shashua1999trajectory,segal20003d,park20113d}. Vo et al.  \cite{vo2016spatiotemporal} used  physics-based motion priors such as least kinetic energy, to formulate a bundle adjustment framework for jointly optimizing  static and dynamic 3D  structure, camera poses and cross-capture temporal offsets.

% \noindent {\bf Low rank constrain}\todo{low-rank NSFM}. 
\noindent {\bf Spatio-temporal smoothness}. Enforcing spatio-temporal smoothness on the geometric estimation process  \cite{park20103d,park20153d,zhu20113d,zhu2015convolutional,valmadre2012general,zheng2015sparse,zheng2017self,vo2016spatiotemporal,simon2014separable,simon2017kronecker} has shown to be an effective approach to leverage temporally dense capture, such as those obtained by multiple video observers. Park et al. \cite{park20103d} parameterize a 3D trajectory in terms of linear combinations of a set of Direct Cosine Transform trajectory bases and optimize for  each coefficient weight. In \cite{park20153d}, Park et al. improve their method by selecting a small number of DCT bases according to N-fold a cross validation method to avoid low reconstructability cases.  Zhu et al. \cite{zhu20113d} improve this result by adding a set of manual keyframes and adding $L_1$-norm regularization to their optimization  to force sparsity on the DCT basis, instead of N-fold cross validation. Valmadre et al. \cite{valmadre2012general} modify the reconstructability analysis for the trajectory basis solutions and propose two solutions: reducing trajectory bases by setting a gain threshold and applying a high-pass filter. Zheng et al. \cite{zheng2017self,zheng2015sparse} reconstructed dynamic 3D structure observed by multiple unsynchronized cameras with partial sequencing information, by assuming a self-expressive motion prior and implementing a bi-convex optimization problem. Recent works  explicitly model and solve for relationships among dynamic 3D estimates and their spatio-temporal data associations  \cite{agudo2018deformable, agudo2018scalable, agudo2019robust, agudo2020segmentation}. 
Along these lines, Xu et al. \cite{Xu_2019_ICCV} used a graph-based formulation  jointly estimating dynamic 3D structure and its corresponding discrete Laplace operator to reduce reliance on the temporal density and uniformity of the input data.

%-------------------------------------------------------------------------
\subsection{Learning for Sparse Dynamic 3D Geometry}
%Classical convolutional neural network for 2D images assumes a structured 2D grid feature as input. For solving 3D computer vision problems using a convolutional nerual network, there are methods first organize the input 3D points, then apply CNN on the structured data. Some other directly proposed methods can implement 3D convolutions on unstructured 3D points.

\noindent {\bf Structured 3D Data Representations}. Relevant to our problem, some early CNN-based approaches to 3D processing \cite{gupta2014learning, long2015fully} map the 3D representations onto a 2D space, where  traditional CNN machinery is deployed. Such representations forgo  accurate modeling of the  geometric relationships lost or warped during projection. Performing 3D convolutions on volumetric representations \cite{graham2017submanifold, maturana2015voxnet, qi2016volumetric, riegler2017octnet, wu20153d}
encodes 3D positional information and adjacency relations, but may  quantize the representation space, leading alternatively,  to data merging or sparsity.   Riegler et al. \cite{riegler2017octnet} addressed this limitation by implementing 3D convolutions on data organized on an oct-tree data structure. 

\noindent {\bf Unstructured 3D Data Representations}.  Qi et al. \cite{Qi_2017_CVPR}  worked  on unstructured data  enforcing network invariance to different permutations of the input feature by aggregating global information through max pooling. PointNet++ \cite{qi2017pointnet++} improved performance by capturing local structure information. Wang et al. \cite{wang2018deep} propose a continuous convolutional neural network, which similarly to 2D convolutions, computes feature maps in terms of weighted sums of the input features. The use of a multi-layer perceptron (MLP)  enabled  adaptive weight determination based on geometric similarity. Boulch \cite{boulch2019generalizing} computes a denser weighting function which takes into account the entire kernel.

\noindent {\bf Deep learning for Dynamic 3D reconstruction}. Recently, network architectures have been proposed for the NRSfM problem. Kong et al. \cite{kong2019deep, kong2020deep} propose an unsupervised auto-encoder neural network to solve NRSfM problem under an orthogonal camera model by relying on a multi-layer sparse coding framework assumption. Wang et al. \cite{wang2020deep} developed a similar multi-layer sparse coding framework with improved generalization to  weak and strong perspective camera models, along with increased robustness to missing data. Novotny et al. \cite{novotny2019c3dpo} learned a deep network to unambiguously factorize 3D structure and viewpoints  by forcing consistency via canonicalization. 
 Bai et al. proposed an end-to-end deep network  \cite{Bai_2020_CVPR} targeting multi-view %\todo{What is the relation to NRSfM??}
 3D facial reconstruction.  Another unsupervised end-to-end deep network \cite{Sidhu2020}  is introduced by Sidhu, which proposes the first dense neural NRSfM approach.
%------------------------------------------------------------------------
\section{Generalized Trajectory Triangulation}

The goal of  generalized  trajectory triangulation (GTT)  is to recover time-varying 3D structure from  a set of 2D observations with known imaging geometry, but absent of global sequencing relations among input capture frames. Accordingly, GTT may be deemed a structure-only variation of the general non-rigid  structure from motion problem (NRSfM). 
%-------------------------------------------------------------------------
\noindent\textbf{A graph-theoretic formulation}.  \label{Sec:DLOE}
A structure-motion graph representation has been recently presented in \cite{Xu_2019_ICCV}, where nodes are mapped from input images and have 3D geometry as attributes, while edges have the pairwise affinities as weights. Based on this representation, the GTT problem can be formulated as jointly estimating dynamic 3D geometry with the graph’s Laplacian matrix, given by 
%Thus, the authors define a tri-convex optimization framework that encodes spatio-temporal priors, including anisotropic smoothness, topological compactness, and multi-view reconstructability. 
%The joint estimation of 3D structure $\mathbb{X}$ and  temporal relations among frames has been recently formulated  in terms of a  discrete Laplace operator (DLO) estimation instance . nThe DLO for a weighted {\em undirected} graph $G = (V,E)$, is defined in terms of the Laplacian  matrix:
\begin{equation}
\mathbb{L} = diag(\mathbb{A} \cdot \mathbf{1}  ) - \mathbb{A} \label{Laplace}
\end{equation}
where $\mathbb{A}$ is the graph's  affinity matrix, whose values $\mathbb{A}_{ij}$ correspond to the  edge weights $e_{ij} \in \mathbb{R}_{\geq0}$, characterizing the spatio-temporal relationships among 3D estimates $\mathbb{X}$. This  generalization of the self-expressive motion prior \cite{zheng2017self}, yields  a non-convex optimization problem of the form
\begin{equation}
\underset{\mathbb{X},\mathbb{L}}{\text{min}} \quad \mathcal{S} \left(\mathbb{L} \mathbb{X}\right)+ \mathcal{T} \left(\mathbb{X}^\top\mathbb{L}\mathbb{X}\right) + \mathcal{R}\left(\mathbb L, \Theta  \right) + \mathcal{O}\left(\mathbb X, \Theta  \right),
\label{LapOptOrigXL}
\end{equation}
where 
%$\mathbb{L}$ denotes the motion graph Laplacian, $\mathbb{X}$ the Euclidean structure matrix, and 
$\Theta=\{  \{\mathbf{x}_{np}\},\{\mathbf{K}_n\},\{\mathbf{M}_n\} \}$ denotes the aggregation of all input 2D observations and their camera parameters, $\mathcal O(\cdot)$ is a data term based on reprojection error, while $\mathcal S(\cdot)$, $\mathcal T(\cdot)$, and $\mathcal R(\cdot)$, are regularizers controlling, respectively, anisotropic smoothness, topological compactness, and multi-view reconstructability.
Variables $\mathbb{X}$ and $\mathbb{L}$  are solved alternatively. That is, for fixed $\mathbb{L}$,  3D structure $\mathbb{X}$ is estimated by unconstrained quadratic programming; while for fixed $\mathbb{X}$, $\mathbb{L}$ is estimated by a linearly constrained quadratic problem. We refer readers to the original publication for further details \cite{Xu_2019_ICCV}.
While the above formulation achieved state of the art 
%results in terms of 
accuracy and robustness, its explicit full graph analysis limits its computational scalability.  GTT-Net aims to alleviate this limitation by developing an encoder-decoder framework  directly mapping the input 3D geometry $\mathbb X$ to the discrete Lapalace operator $\mathbb L$.  
%-------------------------------------------------------------------------
%------------------------------------------------------------------------
\section{GTT-Net}

\begin{figure}
    \begin{subfigure}{0.5\textwidth}
        \centering
        \includegraphics[scale=0.25,trim={0cm 3.3cm 0cm 1.5cm},clip]{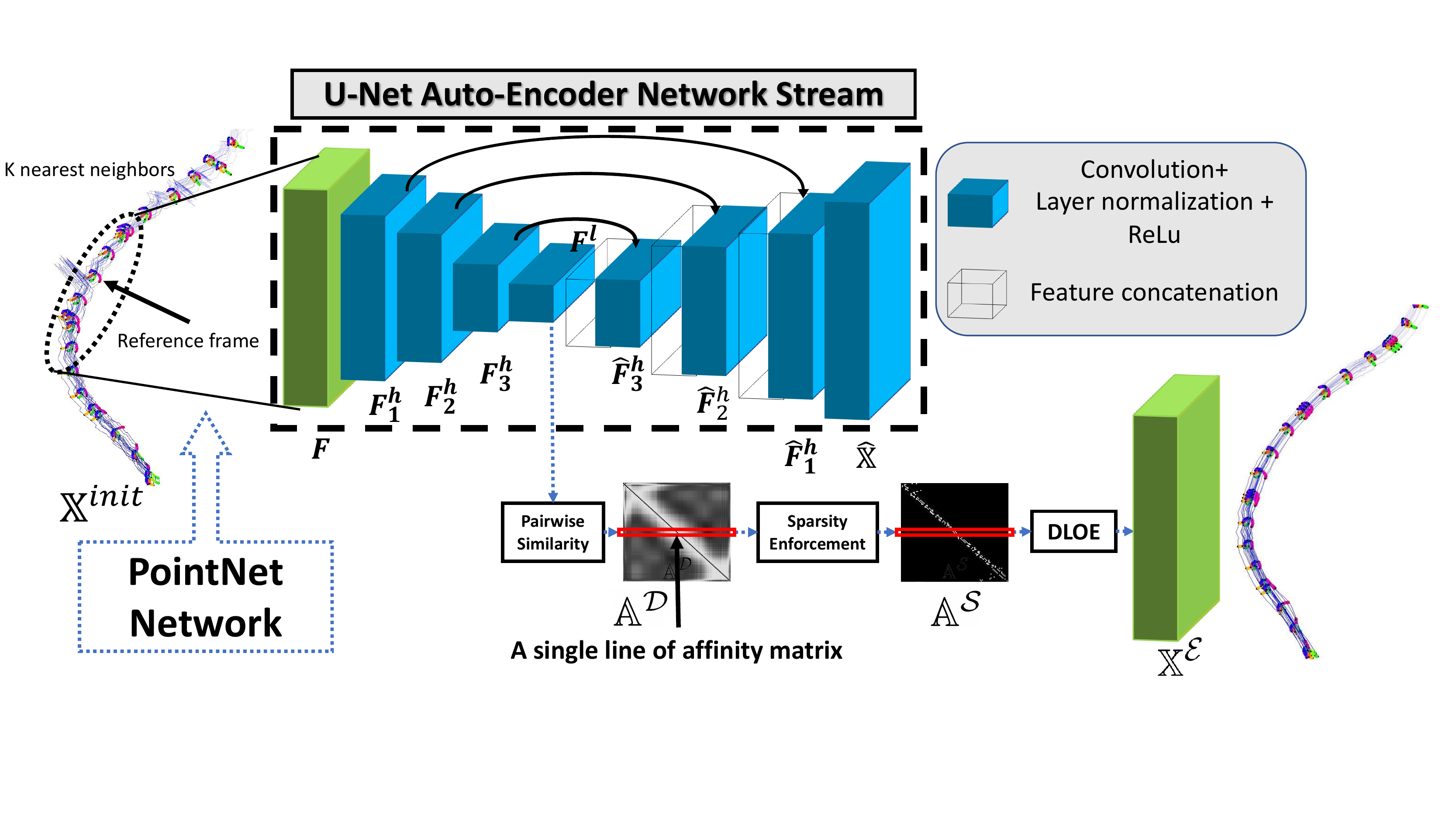}
        \caption{GTT-Net architecture}
        \label{fig:network}
    \end{subfigure}
    \begin{subfigure}{0.5\textwidth}
        \centering
        \includegraphics[scale=0.23,trim={0cm 2cm 0cm 1.2cm},clip]{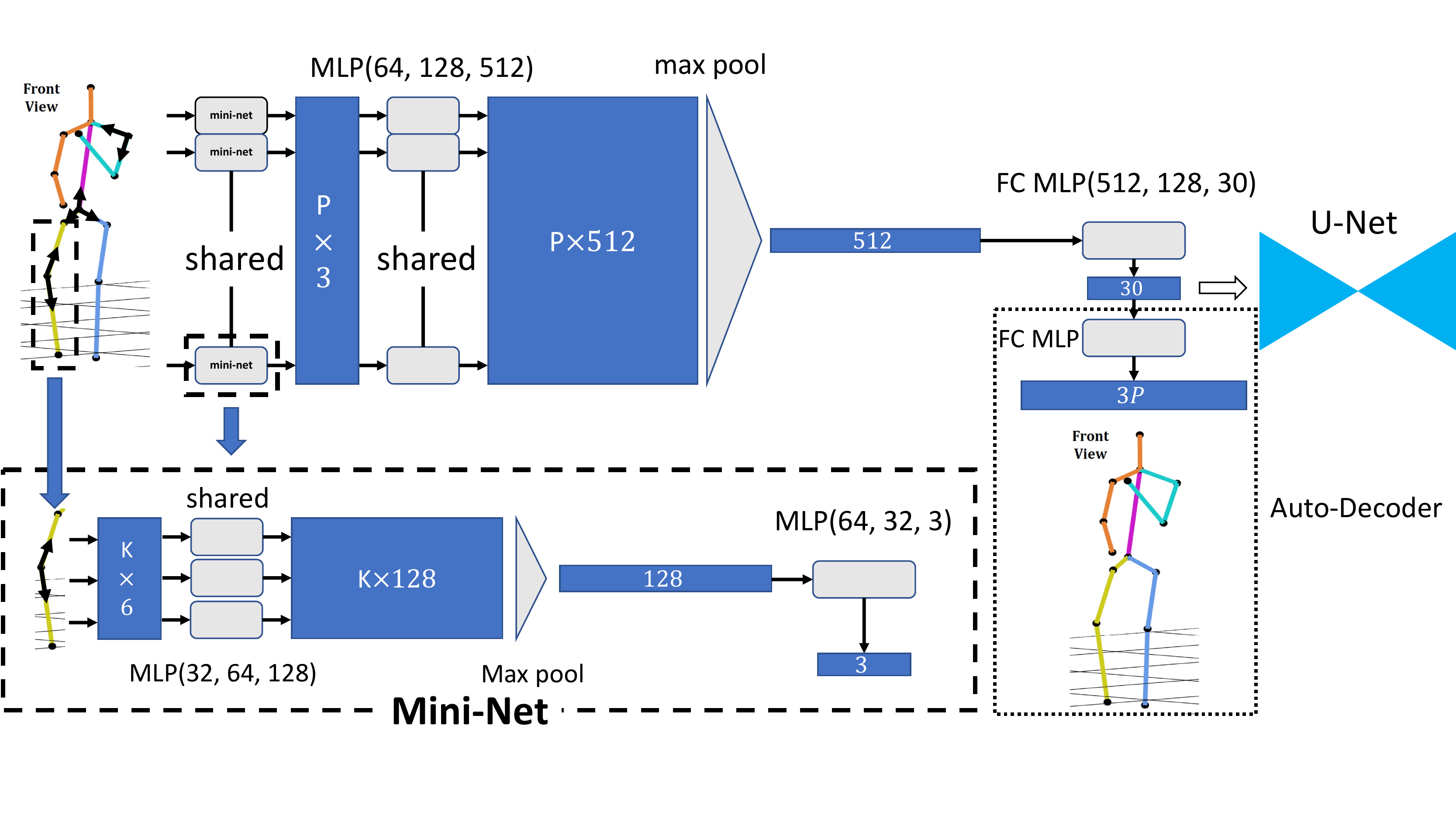}
        \caption{PointNet Auto-encoder}
        \label{fig:pointnet}
    \end{subfigure}
    \caption{(a) GTT-Net combines a U-Net learning a latent space from  input features and  affinity learning layers decoding a pairwise affinity values. (b) An optional PointNet auto-encoder maps input 3D shape structure to an abstract representation with fixed dimensionality. }
\end{figure}

%------------------------------------------------------------------------
As presented in \cite{Xu_2019_ICCV}, 
%geometry estimation is computationally efficient (i.e. independent, local and massively parallel). However, 
 global dependencies required for affinity matrix optimization impose a computational bottleneck. GTT-Net learns to directly estimate these affinity values from input data. 
From an initial  geometry   $\mathbb{X}^{init}$,
we  learn a 
%reduced-dimensionality 
latent space $F^l$  encoding the affinity among input 3D shapes. A  sparse affinity matrix $\mathbb{A}^\mathcal{S}$ decoded from this latent space is fed to a differentiable quadratic optimization module to determine a refined dynamic geometry estimate $\mathbb{X}^{\mathcal{E}}$.
We use data augmentation to explicitly target  equivariance w.r.t. relevant input capture variants and perturbations. To accelerate training, we utilize cascaded training leveraging supervisory loss functions of increasing complexity.

\subsection{Network Architecture}

\noindent\textbf{Parameterizing  Input Geometry}. 
A time-varying set of $P$ 3D points ${\mathbf{X}_{np}}$ is observed  in $N$ images ${\mathcal I_n}$ captured by unsynchronized perspective cameras with known intrinsic and extrinsic matrices $\mathbf{K}_n$ and $\mathbf{M}_n$. 3D points are denoted as $\mathbf X_{np}$, while their image  projections are $\mathbf x_{np}$. The set of all 3D points to estimate is represented by a $N \times 3P$ matrix
%All observation are The 3D points across all images are presented as a $N \times 3P$ matrix
\begin{equation}
\mathbb{X} = 
\begin{bmatrix}
    \mathbf{X}_{11} & \dots & \mathbf{X}_{1P}  \\
    \vdots & \ddots &\vdots \\
    \mathbf{X}_{N1} & \dots & \mathbf{X}_{NP}  
\end{bmatrix}
\end{equation} 
where $\mathbf X_{np}$ represents a 3D point's coordinates. 
%corresponding to the 2D observations $\mathbf x_{np}$. 
Each row of $\mathbb{X}$ 
%represents the $3P$ dimensional feature corresponding to 
aggregates
the $P$ 3D points captured in frame $n$ and  constitutes a per-frame shape descriptor from which to estimate affinities.
%,  our basic unit of geometric analysis. 
The input matrix $\mathbb X^{init} $ 
%will be the input to our network, after being
is estimated through pseudo-triangulation of viewing rays associated with $\mathbf x_{np}$. %\todo{Matrix $\mathbb X $should be transposed??}

\begin{figure}[t!]
        \centering
        \includegraphics[scale=0.26]{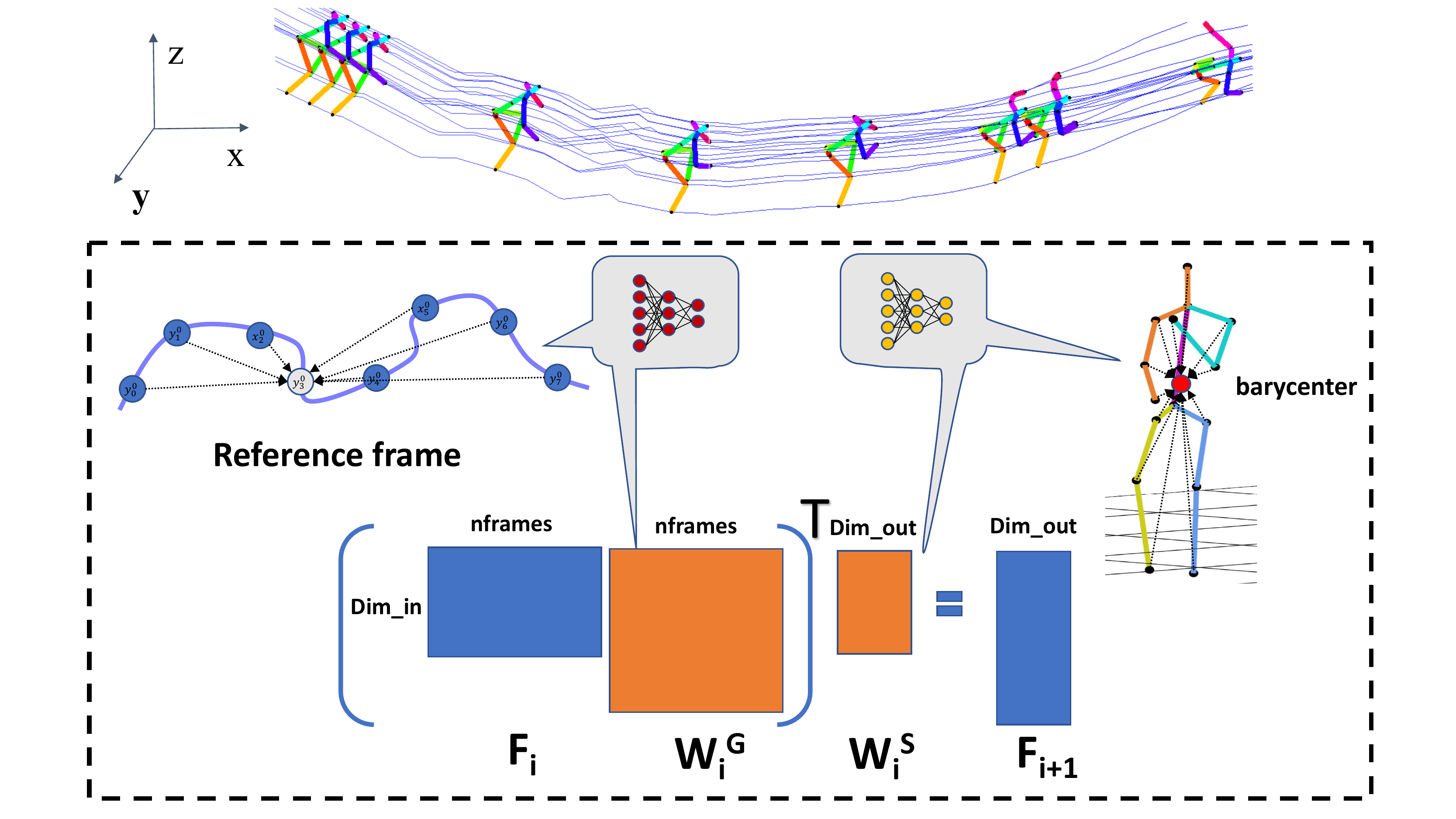}
        \caption{Each U-Net layer learns two types of continuous convolution filters: One applied among shape descriptors along the entire  motion trajectory $(W_i^G)$ and another based on intra-shape 3D point geometry ($W_i^S$). }
        \label{fig:conv3D}
\end{figure}
\noindent\textbf{Parametric Continuous Convolution Layers}. 
\label{Sec:3D_conv}
Based on \cite{wang2018deep} and \cite{boulch2019generalizing}, we perform approximated continuous convolution operations on a given feature descriptor $x$ as
\begin{equation}
    h(x) = \int_{-\infty}^{+\infty}f(y)g(x-y)dy \approx \sum_{j \in \mathcal{N}^x_K} ^K\frac{1}{K}f(y_j)g(x-y_j)
\end{equation}
where $y_j$ is one the $K$ nearest  neighbors 
%the reference frame 
of $x$, 
%sampled from the support domain, 
$f$ is the feature map value function 
%(which in our case corresponds %to the 3D coordinate of all $P$ joints the same as the support domain   \todo{... 
%directly to the feature map value) 
and $g$ is a convolution kernel function approximated by a multi-layer perceptron (MLP) 
\begin{equation}
    g(x-y_j;\theta) = MLP(x-y_j;\theta).
    \label{Eq:Conve}
\end{equation}
This yields continuous output values 
%along the entire support domain 
using a finite set of learned weight parameters $\theta$. We learn two types of filters for each layer, see Fig. \ref{fig:conv3D}.  
The first operates on the N single-frame descriptors and their K nearest neighbors, defining the support neighborhood w.r.t. spatio-temporal proximity among their shapes. Filter values  are determined by geometry difference between shapes according to Eq. \ref{Eq:Conve}.
The second operates on single-coordinate whole-trajectory descriptors and defines the support neighborhood domain w.r.t. intra-shape geometry (i.e.  per-component proximity to their barycenter). Filter values are determined by geometry difference between joints.
%Intuitively, the first type uses the rows of $\mathbb X$ as the support domain, while the second uses its columns

%we first see all the 3D joints in a frames together as a high-dimension point and K nearest neighbors as a trajectory. The parameters of the first filter are computed based the geometry difference to the reference frame and applied on this high-dimension trajectory. Then, we see a joint along all K neighbors as a high-dimension joint and the parameter of the second filter are computed based the geometry difference to the barycenter and applied on this high-dimension structure.

\noindent\textbf{U-Net Auto-Encoder Network Stream}. 
We learn a latent space $\mathbf{F}^{l}$  using a U-Net  encoder-decoder  to perform dimensionality reduction through continuous parametric convolutions, see  Fig. \ref{fig:network}. 
For translation and scale invariance across different input data, we apply layer normalization \cite{ba2016layer} for  input and hidden layers by subtracting the mean $\mu$, dividing by the standard deviation $\sigma$ for each feature channel, while scaling and shifting  by learnable parameters $\gamma$ and $\beta$.%\todo{is this per feature channel?: yes, maybe we should put the equation here?}
\begin{equation}
    \hat{x}_{d,i} = \frac{x_{d,j}-\mu_d}{\sqrt{\sigma_d+\epsilon}}\gamma_d +\beta_d
\end{equation}
An affinity matrix $\mathbb{A}^\mathcal{D}$ is computed in closed form as the pairwise similarity between latent space features  $\mathbf{F}^{l}$ by 
%To ensure $\mathbb{A}^\mathcal{D}_{ij} \geq 0$, we apply a fixed bias term $b=1$, yielding affinity values  ranging from 0 to 2. 
%This matrix explicitly encodes the pairwise affinities among per-frame shape descriptors in our input. 
\begin{equation}
    \mathbb{A}^\mathcal{D}_{nm} = \frac{1}{(1+exp||\mathbf{F}_n^l - \mathbf{F}_m^l||)}
\end{equation}
Unlike a regular graph affinity matrix, $\mathbb{A}^\mathcal{D}$ does not encode a graph's local connectivity.  $\mathbb{A}^\mathcal{D}$ is  sparsified into $\mathbb{A}^\mathcal{S}$ through a %subsequent 
layer retaining the $Q$-highest  affinity values among the per-feature convolution support domain $\mathcal{N}_K^x$. Empirically, we found $Q$=2 yielded the best performance (see Fig. \ref{fig:N affinity}) and enforced this selection criteria deterministically. 
% shows how keeping the top two affinity values yield the best results.
Finally, $\mathbb{A}^{\mathcal{S}}$ is fed into a differentiable instance of the Discrete Laplace Operator Estimator framework \cite{Xu_2019_ICCV}, denoted as a DLOE-layer, to estimate output 3D geometry  $\mathbb{X}^{\mathcal{E}}$.

\noindent\textbf{PointNet Network Stream}. 
To allow for input shapes having different number of 3D points, we integrate a PointNet network \cite{Qi_2017_CVPR} to provide a fixed-sized input into our U-Net, see Fig. \ref{fig:pointnet},
%which will map any structure to a fixed number of virtual key points. 
We normalize each shape by subtracting its barycenter before PointNet  maps it to a 30 dimension feature. To retain spatial separation among shapes we interpret PointNet's output as 10 virtual 3D points, add back the original barycenter, and feed them to the U-Net.  %Each class of structure will have to train it's own Auto-decoder to reconstruct back to original structure.

\subsection{Supervisory Data }
\label{supervisory}

%Training leverages ground truth geometry  from multi-view 3D motion capture. 
%Known 3D structure $\mathbb{X}^{\mathcal{G}}$ and  sequencing information can be used to supervise our training.
%determine a ground-truth proxy affinity matrix $\mathbb{A}^{\mathcal{G}}$.

\begin{figure}[t]
        \centering
        \includegraphics[scale=0.3,trim={0cm 0cm 0cm 0cm},clip]{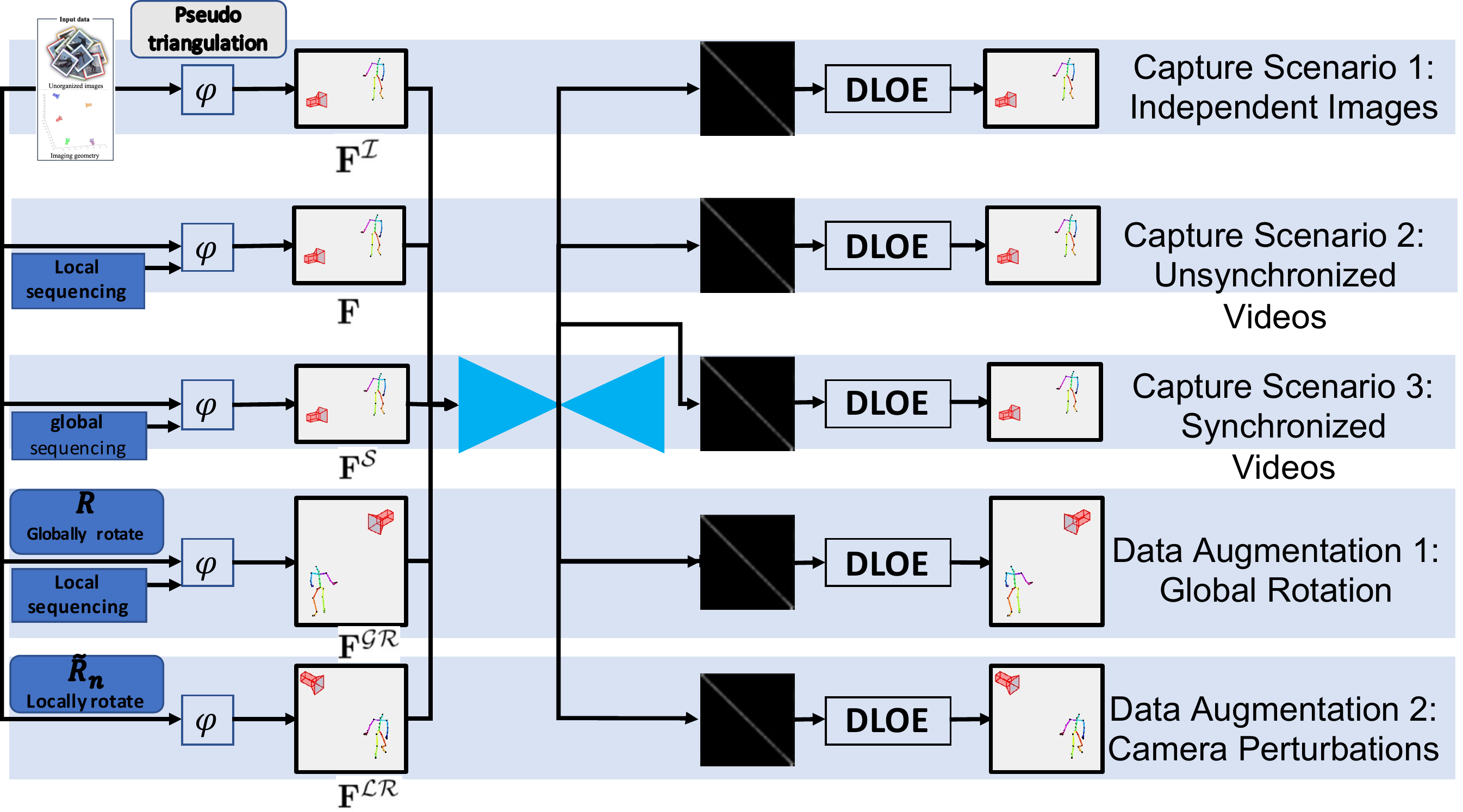}
        \caption{Five different variants of the input feature are generated and the network trained with shared weights. The convolution support domain is determined independently for each input variant.}
        \label{fig:varints}
\end{figure}

%Motion capture 3D values are used to synthesize different capture scenarios varying temporal sampling density and viewing geometry. 
Depending on the capture scenario, complete or partial sequencing priors (e.g. sequencing among frames belonging to the same camera or video stream) may be available. As GTT-Net encodes these priors in terms of the support domain $\mathcal{N}_K^x$ used for continuous convolution, we only need to train a single network instance that is inclusive of all such variations. 
We explicitly instantiate such input prior variations  within our training data and to account for capture variability, we perform data augmentation tailored to our formulation as in Fig. \ref{fig:varints}. We inject Gaussian noise to the 2D features $\mathbf{x}_{np}$  to account for feature localization ambiguity and apply geometric transformations to the ground truth data
to account for capture variability. 
%As motivation, for the same 3D dynamic scene, 2D observations and initial 3D feature input depend on the camera position relative to a global coordinate system.
%During the training, we generate five input variants to simulate the conditions of those changes.

%For training we leve During training the proposed network, for each sequence of 3D datasets, we will generate a few variants of input feature based on different sequencing information given as in Fig. \ref{fig:varints}, then we enforce the network outputs consistent result which should all be closed to the pseudo ground true affinity matrix $\mathbb{A}^{\mathcal{G}}$ and the ground truth 3D structure $\mathbb{X}^{\mathcal{G}}$, so the network could be robust for different sequencing information inputs.

\noindent {\bf Capture Scenario 1: Independent Images}.
Independent imagery provides no sequencing information.
%and constitute the default setup for GTT-Net.
The convolution support domain for shape descriptors is determined by the spatial distribution of our initial 3D geometry  $\mathbb{X}^{init}$, which is computed by exhaustive pseudo-triangulation of sparse 2D features. Once a rough 3D geometry is estimated per each frame, we  compute the per-frame K-nearest neighbors 
by the combination of triangulation error and viewing ray convergence analysis to eliminate frames with reduced  camera baseline and unreliable triangulation.
This input feature variant is denoted as $\mathbf{F}^{\mathcal{I}}$.

\noindent {\bf Capture Scenario 2: Unsynchronized Videos}.
For unsynchronized videos, sequencing priors are available  for each independent video stream, allowing us to summarily eliminate from the  support domain any frames from the same stream, and frames within another stream that are not adjacent among themselves. These constraints mitigate repetitive and/or self-intersecting 3D motions.  
 The initialized input feature is defined as $\mathbf{F}$.
 %=\mathbb{X}^{init}$.  

\noindent {\bf Capture Scenario 3: Synchronized Videos}. 
For synchronized videos,\footnote{Synchronization denotes temporal alignment, not capture concurrency}  global sequencing is known and we can readily determine the K-nearest neighbors as  those elements temporally adjacent to a given reference video frame. Also, pseudo-triangulation efficiency and reliability can benefit from guidance from the known sequencing info.
%Also, 3D structure initialization are computed by only comparing the triangulation error with two nearby frames.
This input feature variant is denoted as $\mathbf{F}^{\mathcal{S}}$.

%\subsection{Data Augmentation}

\noindent {\bf Data Augmentation 1: Global Structure Rotation}.
Feature normalization in our encoder layers mitigates global scale and displacement variations. To promote rotational invariance we generate augmented input instances by randomly rotating initial 3D structure and camera poses jointly. While this transformation does not change input 2D feature locations, it targets the generalization of 3D and sequencing estimates.  This input feature variant is denoted as $\mathbf{F}^{\mathcal{GR}}$.

\noindent {\bf Data Augmentation 2: Camera Perturbations}.
%Based on the fact that the camera poses can be anywhere in the real situation specially if all the images are captured independently. We hope that for the same scene that our network can be invariant for any camera poses. 
We inject structured perturbations to our input by randomly rotating and translating the camera pose of each frame. Since this transformation changes the imaging geometry, it alters the input 2D features used to initialize both 3D structure, and the K nearest neighbors associated to each frame. This input feature variant is denoted as $\mathbf{F}^{\mathcal{LR}}$.

\subsection{Loss functions}

\noindent \textbf{U-Net Reconstruction Loss}. 
To train our U-Net auto-encoder, we penalize the difference between the  input and the reconstructed output maps, which correspond, respectively, to the initialized 3D structure $\mathbb{X}^{init}$ and a decoded 3D structure $\hat{\mathbb{X}}$. We penalize the differences between each hidden feature map $\mathbf{F}_i^h$ inside the encoder and the symmetrically corresponding hidden feature map $\hat{\mathbf{F}}_i^h$ in the decoder as in Fig. \ref{fig:network}. The loss function is written as,
\begin{equation}
    \ell^{\text{\AE}} = \frac{1}{NP}(||\mathbb{X}^{init} - \hat{\mathbb{X}}||^2_F + \sum_i^{d-1}||\mathbf{F}_i^h - \hat{\mathbf{F}}_i^h||^2_F),
\end{equation}
where $d=4$ is the depth of the encoder and decoder layers.

\noindent \textbf{(Pseudo) Ground Truth Affinity Loss}. %($\ell^{\mathcal{A}}$)}
Ground truth affinity matrix optimization is computationally intractable (NP-hard). Hence, we use ground truth sequencing  to generate a proxy (pseudo) ground truth affinity matrix $\mathbb{A}^{\mathcal{G}}$ having affinity values $\mathbb{A}^{\mathcal{G}}_{i,j} = 1$ for temporally consecutive frames and  zero otherwise. If ground truth structure is available, we estimate real-valued affinities through optimization as in \cite{Xu_2019_ICCV}. The reconstruction accuracy training by these two kinds of (pseudo) ground truth affinity matrix are compared in Fig. \ref{fig:ablation}.
We penalize the difference between $\mathbb{A}^{\mathcal{D}}$ %\todo{should this be $\mathbb A^s$???} 
and $\mathbb{A}^{\mathcal{G}}$.
\begin{equation}
    \ell^{\mathcal{A}} = \frac{1}{N}\sum_i^{d}||\mathbb{A}^{\mathcal{D}} - \mathbb{A}^{\mathcal{G}}||^2_F
\end{equation}

\noindent \textbf{3D Reconstruction Loss}. % for Supervised Learning}. %($\ell^{\mathcal{X}}$)}
Given the affinity matrix $\mathbb{A}^{\mathcal{S}}$ estimated by GTT-Net, we generate the corresponding Laplacian matrix as in Eq.(\ref{Laplace}) and estimate the 3D geometry $\mathbb{X}^{\mathcal{E}}$ by solving a quadratic programming problem. We penalize the 3D structure estimation error w.r.t.  ground truth  $\mathbb{X}^{\mathcal{G}}$  as
\begin{equation}
    \ell^{\mathcal{X}} = \frac{1}{NP}||\mathbb{X}^{\mathcal{E}} - \mathbb{X}^{\mathcal{G}}||^2_F
\end{equation}

\noindent \textbf{Smoothness Loss}. % for Weakly Supervised Learning}. %($\ell^{\mathcal{S}}$)}
In the absence of ground truth 3D structure $\mathbb{X}^{\mathcal{G}}$, we penalize the first and second terms in Eq.\ref{LapOptOrigXL},  to foster local smoothness and linear topological structure. %\cite{Xu_2019_ICCV}

\begin{equation}
    \ell^{\mathcal{S}} = \mathcal{S} \left(\mathbb{L} \mathbb{X}\right)+ \mathcal{T} \left(\mathbb{X}^\top\mathbb{L}\mathbb{X}\right)
\end{equation}

\noindent \textbf{PointNet Auto-encoder reconstruction Loss}.  %($\ell^{\mathcal{P}}$)}
If the PointNet stream is considered, we  penalize the difference between it's input $\mathbb{X}^{init}$ and output map reconstructed by a domain-specific decoder $\hat{\mathbb{X}}^{P}$. In this scenario, the input to our  U-Net is PointNet's  fixed-dimension latent space.
\begin{equation}
    \ell^{\mathcal{P}} = \frac{1}{NP}(||\mathbb{X}^{init} - \hat{\mathbb{X}}^{P}||^2_F)
\end{equation}

\subsection{A Cascaded Supervision Strategy.}
\begin{figure}[t!] 
\centering
    \includegraphics[scale = 0.25,trim={0cm 0cm 0cm 0cm},clip]{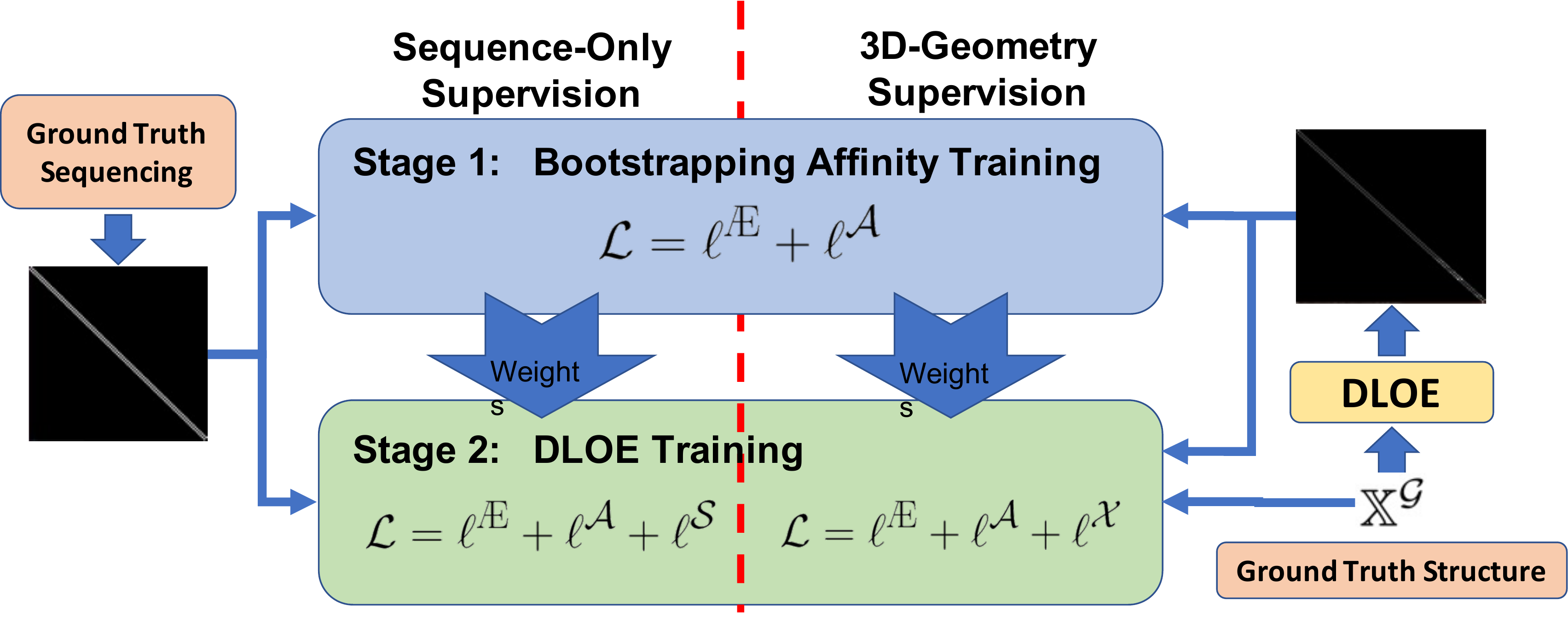}
    \caption{Cascaded Supervision Strategy. }
    \label{fig:cascade}
\end{figure}
The loss functions just described address a diversity of performance aspects we aim to control through supervision. However, they impose different levels supervisory specificity as well as computational burden. In order to streamline the training process, we partition it into sequential stages, each one of them considering supervisory loss functions of increasing specificity and complexity. We aim to bootstrap the training process using efficient weak supervision and later improve upon the quality of the results by incorporating more targeted and computationally burdensome loss functions. %The empirical motivation for this is that
We observed that strong supervision based on the output of the DLOE layers, while being the most effective, significantly slowed down convergence rate and increased the processing time for each epoch.
Accordingly, DLOE-based supervision is used to fine-tune affinity estimation and omitted during initial training epochs.
%the preceding stages have converged.
We now describe our 2-stage cascaded approach, shown in Fig. \ref{fig:cascade}.\\

\noindent {\bf Stage 1: Bootstrap Affinity Supervision}.
Stage 1 only enforces sequencing constraints and relies on the $\ell^{\text{\AE}}$ and $\ell^{\mathcal{A}}$ loss functions. The goal is to accurately learn to auto-encode U-Net's input signal, while effectively learning pairwise affinity. For sequencing-only supervision, the binary version of $\mathbb{A}^{\mathcal{G}}$ is used to target the identification of temporal neighborhoods. Conversely, if ground truth 3D geometry $\mathbb X^{\mathcal G}$ is available, the continuous version of $\mathbb{A}^{\mathcal{G}}$ is used to target fine-grain affinity estimation.

\noindent {\bf Stage 2: DLOE-based Supervision}.
Stage 2 leverages the DLOE model to enforce geometric regularization on the output 3D structure. 
For sequencing-only supervision, we enforce the smoothness loss function $\ell^{\mathcal{S}}$, to  learn affinity values in $\mathbb{A^{\mathcal{S}}}$ yielding smooth 3D trajectories. 
%As discussed in the experiment section,  $\ell^{\mathcal{S}}$ is a surprisingly effective regularizer for 3D geometry.
%The evaluation of the smoothness and topological compactness terms does not require additional supervisory data.
For training instances where $\mathbb{X}^{\mathcal{G}}$ is available, we replace $\ell^{\mathcal{S}}$ with a 3D reconstruction loss  $\ell^{\mathcal{X}}$ for fully supervised learning. 

%Exploiting the full DLOE layer functionality provides the most accurate geometry estimation results at the expense of computational burden. 
\section{Experiments}
\begin{figure}[t!]
    \begin{subfigure}{0.1\textwidth}
        \centering
        \includegraphics[scale=0.2,trim={0cm 0cm 0cm 0cm},clip]{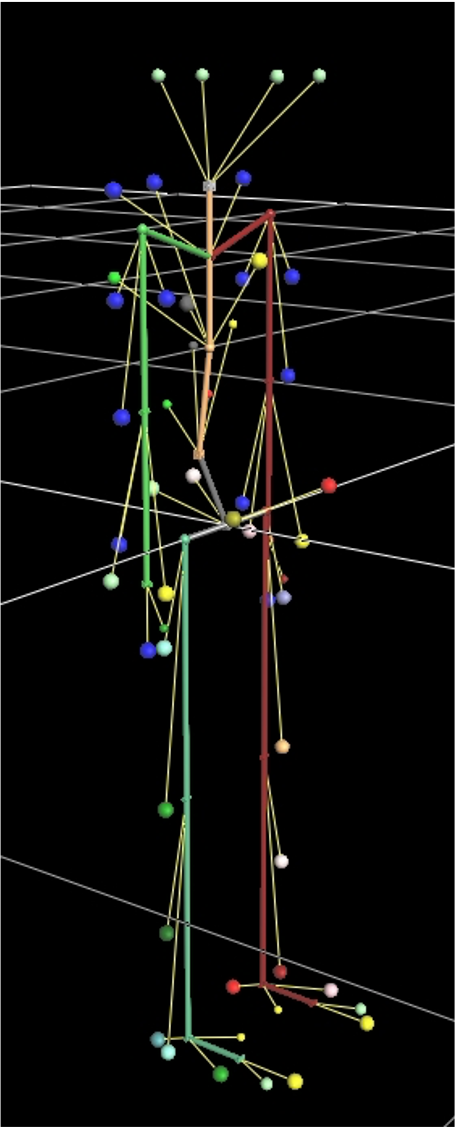}
        \caption{Human}
        \label{fig:CMU_human}
    \end{subfigure}
    \begin{subfigure}{0.2\textwidth}
        \centering
        \includegraphics[scale=0.2,trim={0cm 0cm 0cm 0cm},clip]{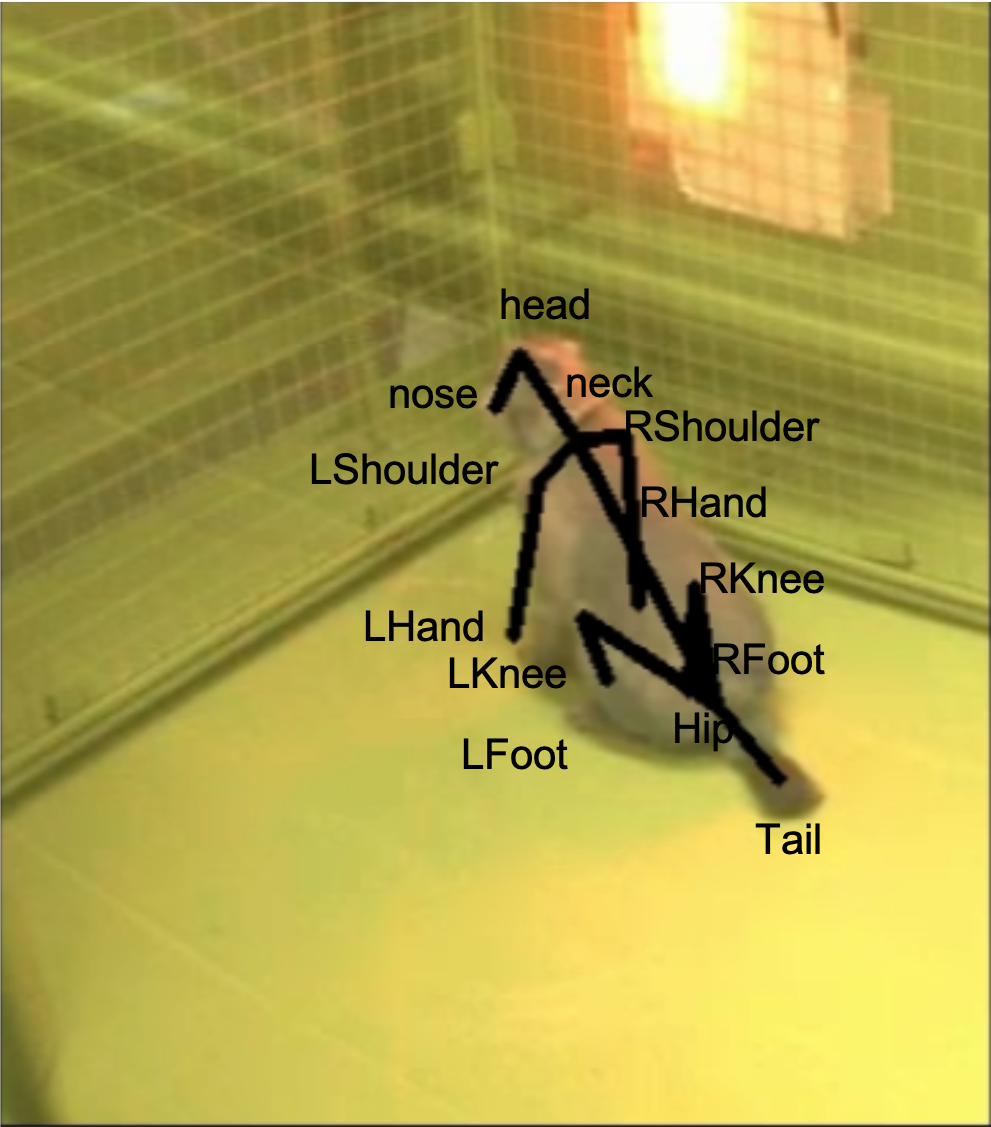}
        \caption{Monkey}
        \label{fig:moneky}
    \end{subfigure}
    \begin{subfigure}{0.1\textwidth}
        \centering
        \includegraphics[scale=0.2,trim={0cm 0cm 0cm 0cm},clip]{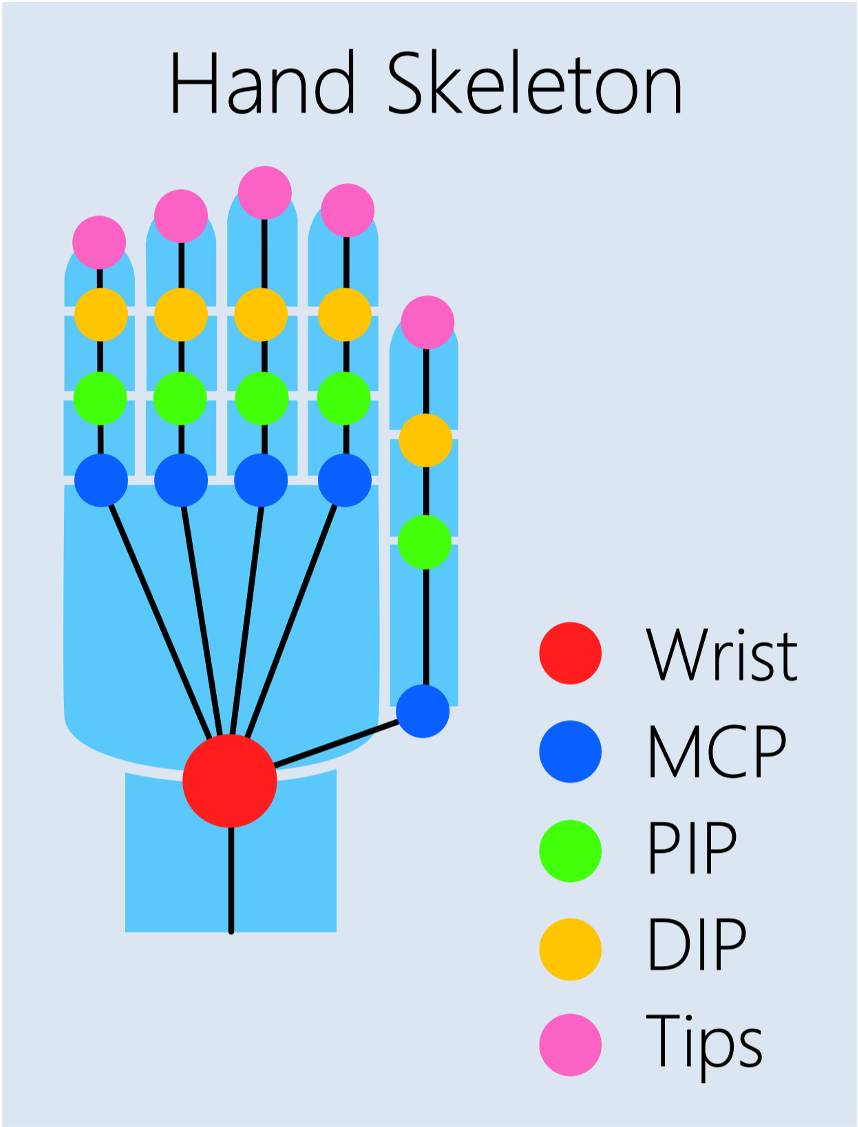}
        \caption{hand}
        \label{fig:hand}
    \end{subfigure}
    \caption{PointNet auto-encoder is trained on three diverse articulated motion datasets: human, monkey and hands. }
    \label{fig:pointnet_train}
\end{figure}

\begin{figure*}[t!] 
\begin{subfigure}[h!] {1.05\textwidth} % width of left 
        \centering
		\includegraphics[scale = 0.5,trim={0cm 0cm 0cm 0cm},clip]{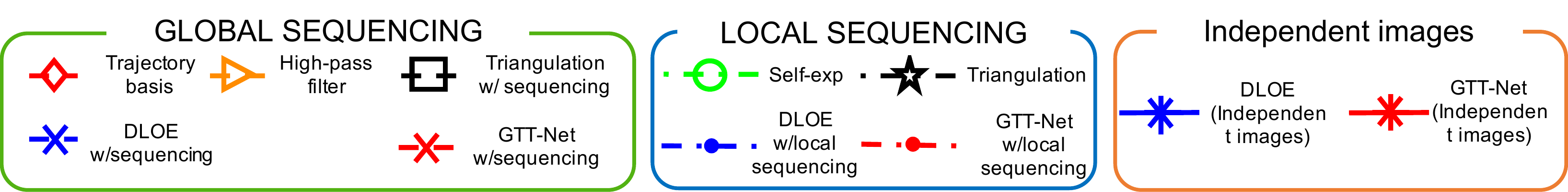}
		\label{fig:legend}
	\end{subfigure}
    \begin{subfigure}[h!] {0.33\textwidth} % width of left 
        \centering
		\includegraphics[scale = 0.155]{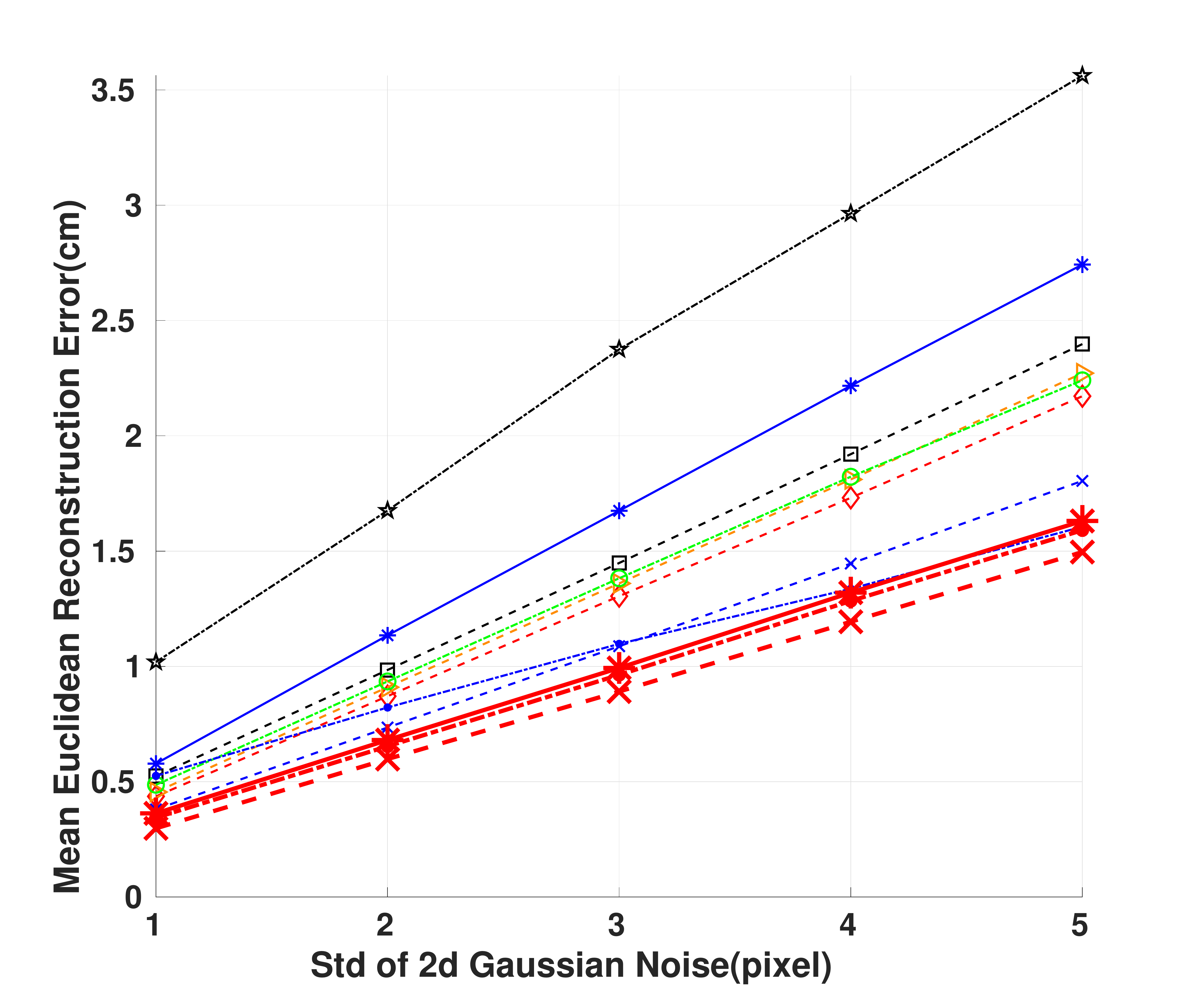}
		\caption{Noise level} 
		\label{fig:DiffNoise}
	\end{subfigure}
	%\hspace{1em}
	\begin{subfigure}[h!] {0.33\textwidth} % width of right
	    \centering
		\includegraphics[scale = 0.155]{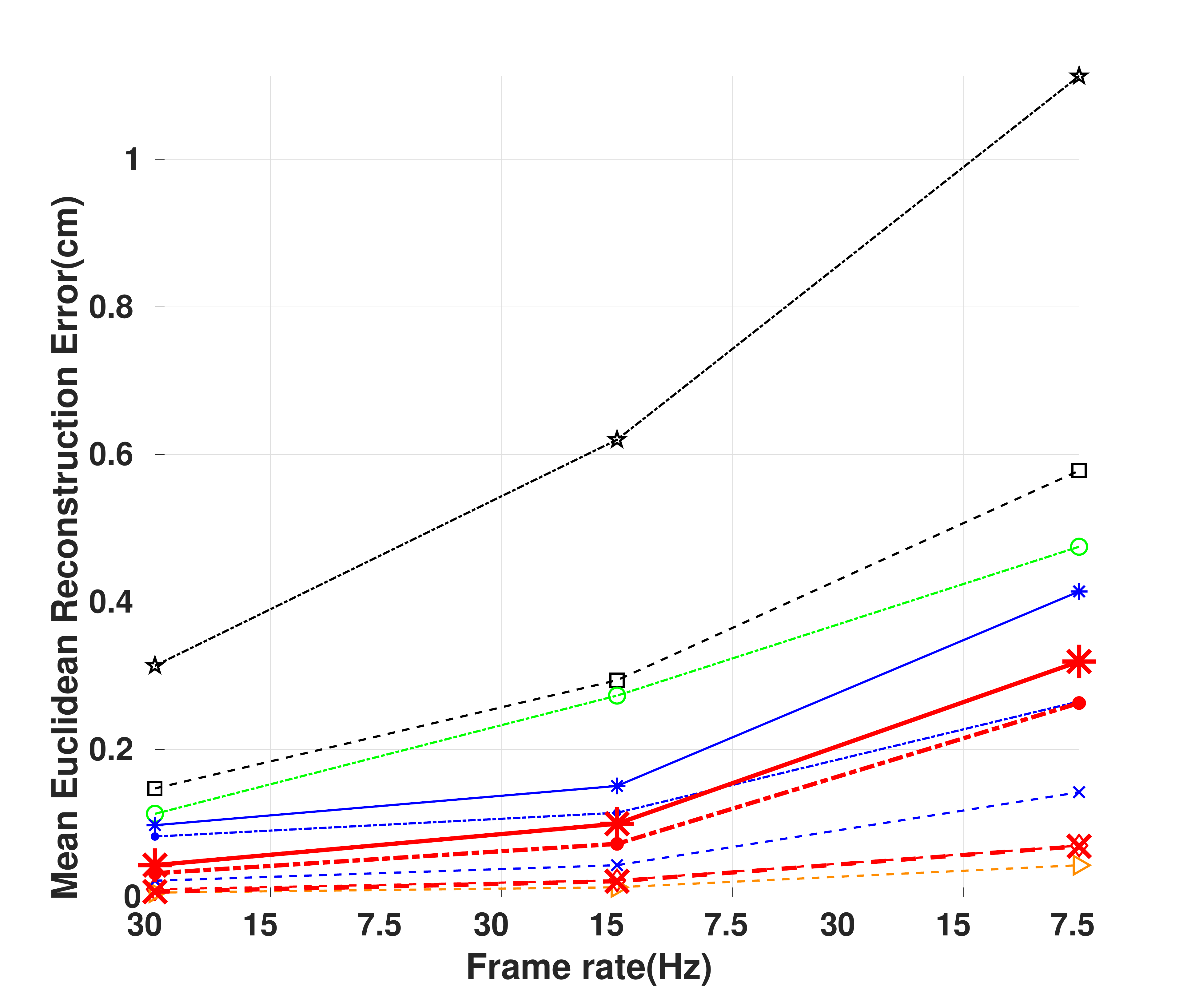}
		\caption{Frame rates} 
		\label{fig:DiffFreq}
	\end{subfigure}
	%\hspace{1em}
	\begin{subfigure}[h!] {0.33\textwidth} % width of right 
	    \centering
		\includegraphics[scale = 0.155]{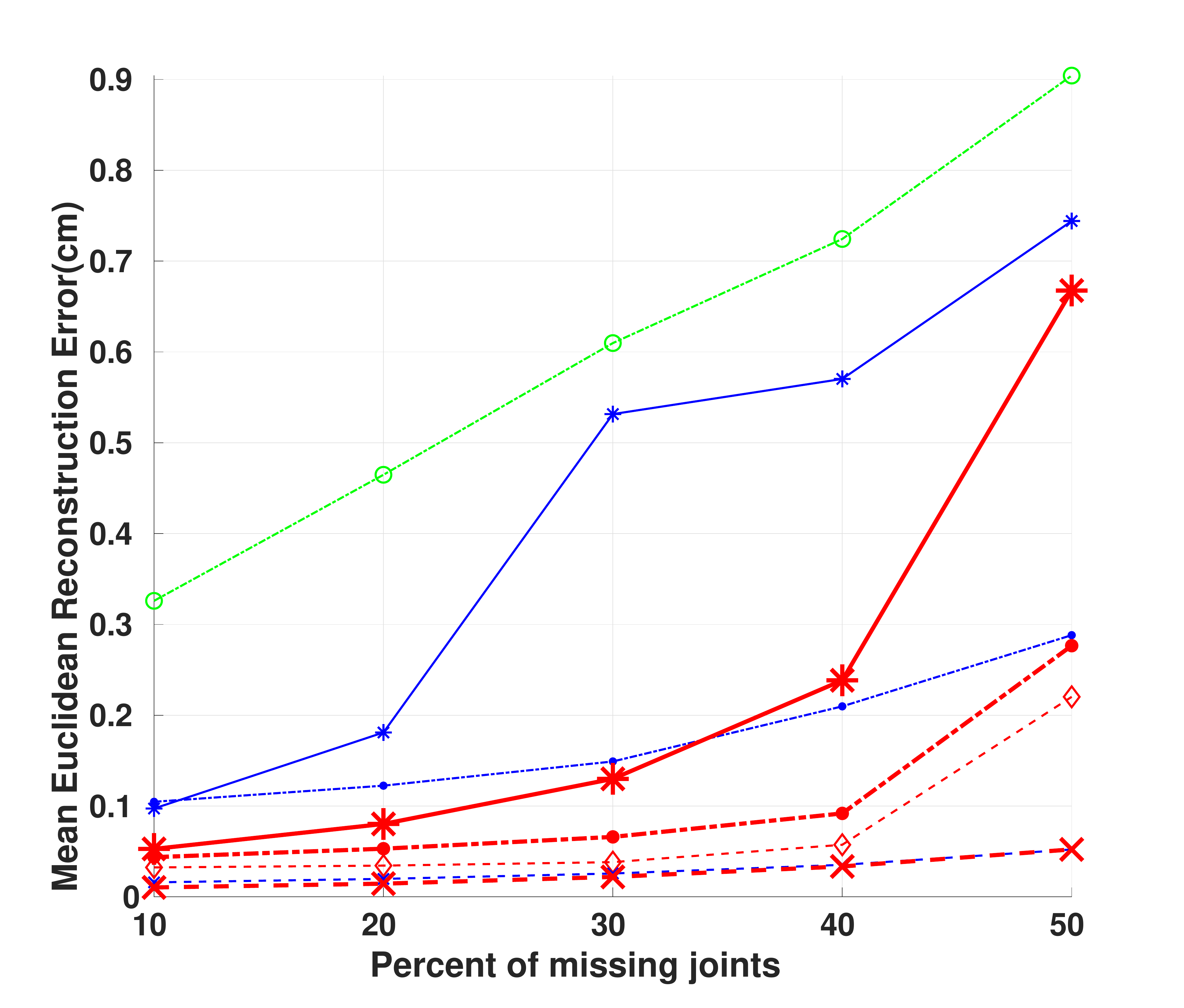}
		\caption{Missing points} 
		\label{fig:DiffMissing}
	\end{subfigure}
	\caption{(a) 3D Reconstruction error of the motion capture datasets under different level of 2D noise, (b) frames rates and (c) different percent of missing frames}
	\label{fig:Accuracy}
\end{figure*}

\begin{figure}[t!]
    \fbox{\begin{subfigure}[t!]{0.22\textwidth}
        \centering
        \includegraphics[scale=0.241,trim={0cm 0cm 0cm 0cm},clip]{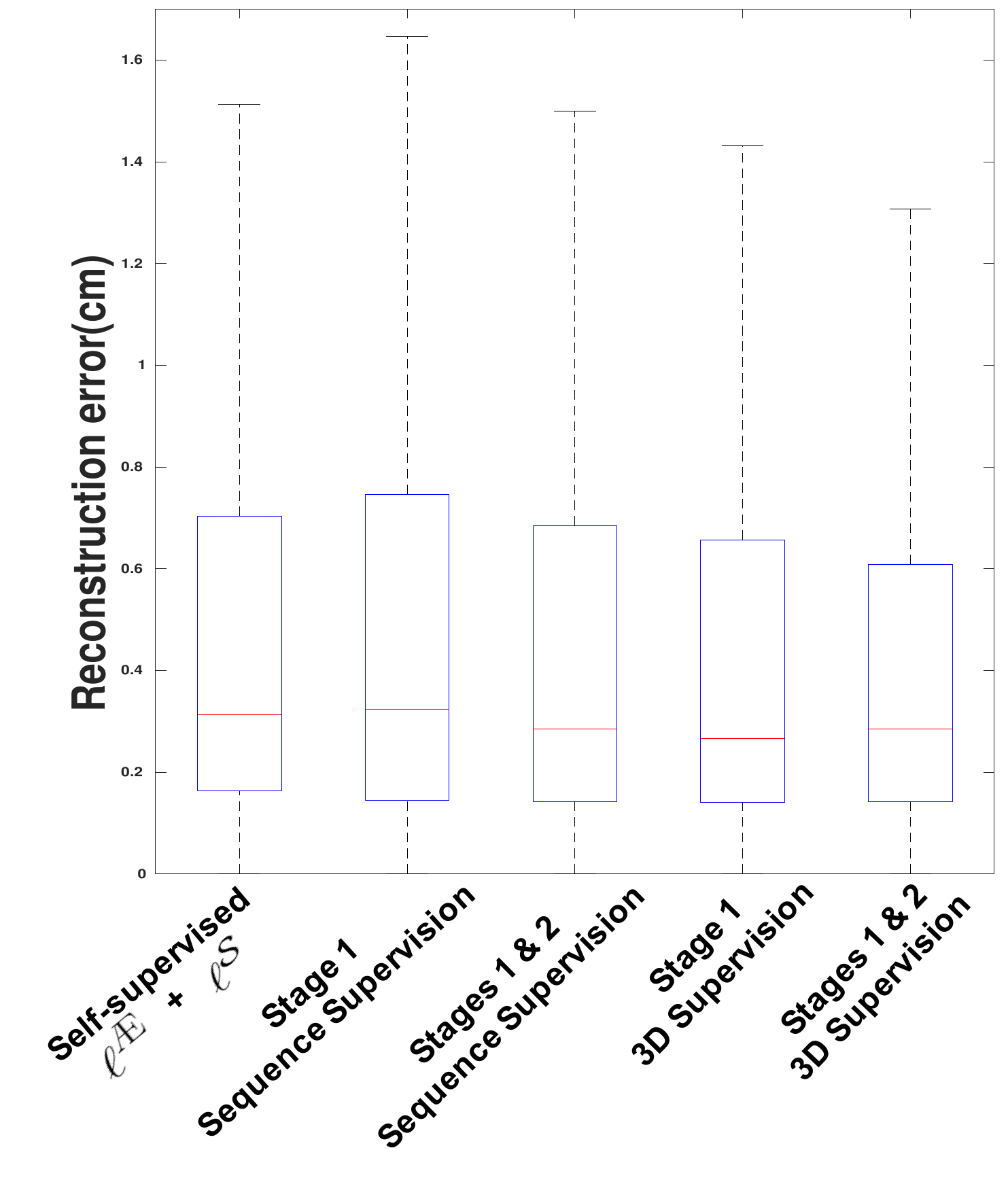}
        \caption{Training ablation}
        \label{fig:ablation}
    \end{subfigure}}
    \fbox{\begin{subfigure}[t!]{0.22\textwidth}
        \centering
        \includegraphics[scale=0.23,trim={0cm 0cm 0cm 0cm},clip]{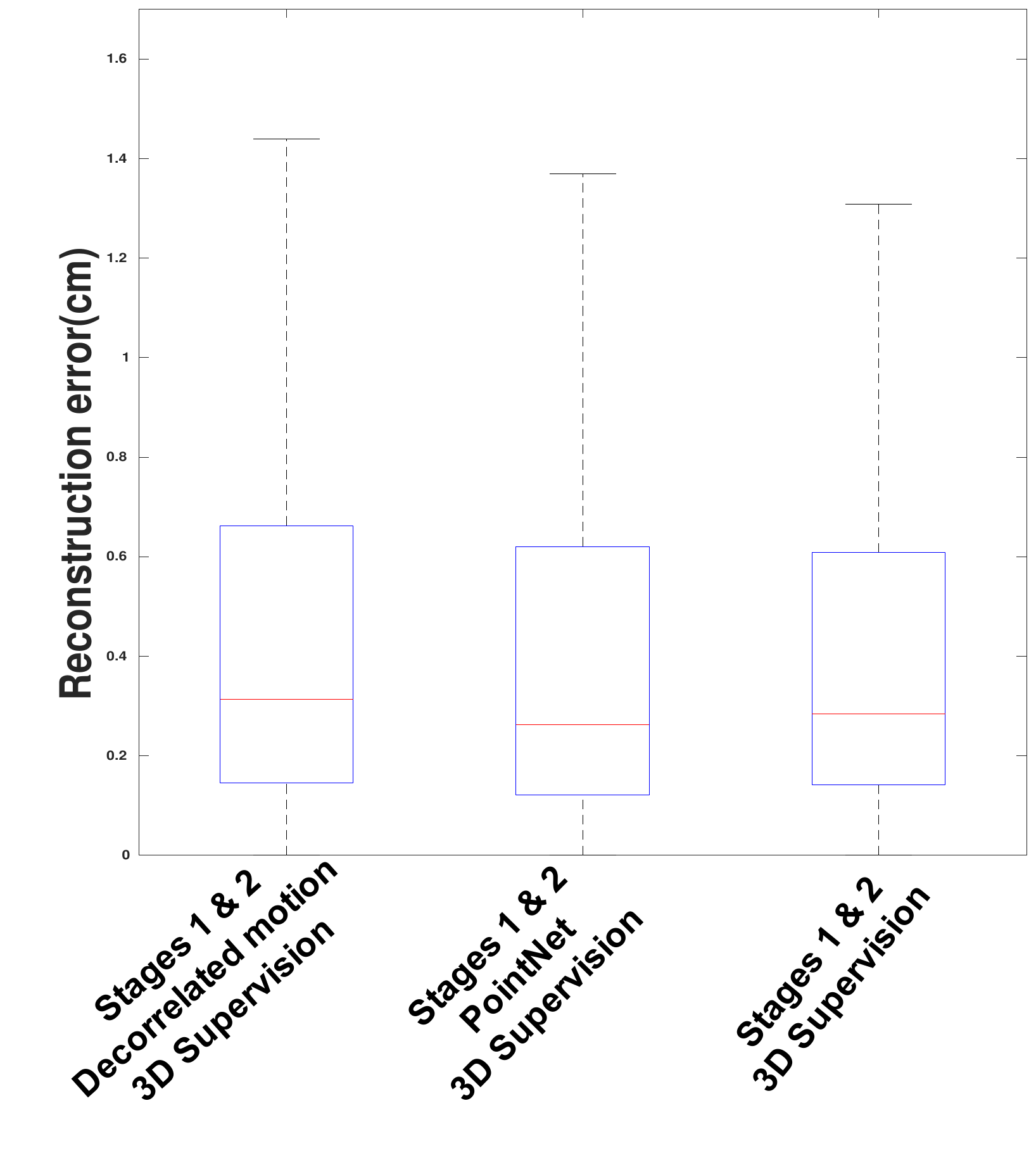}
        \caption{PointNet validation}
        \label{fig:PointNetValid}
    \end{subfigure}}
    \caption{Reconstruction error distributions. (a) Different training cascade variations. (b) Different GTT-Net variants.}
\end{figure}

\begin{figure}[t!]
    \fbox{\begin{subfigure}[t!]{0.24\textwidth}
        \centering
        \includegraphics[scale=0.058,trim={0.2cm 0cm 0.7cm 0cm},clip]{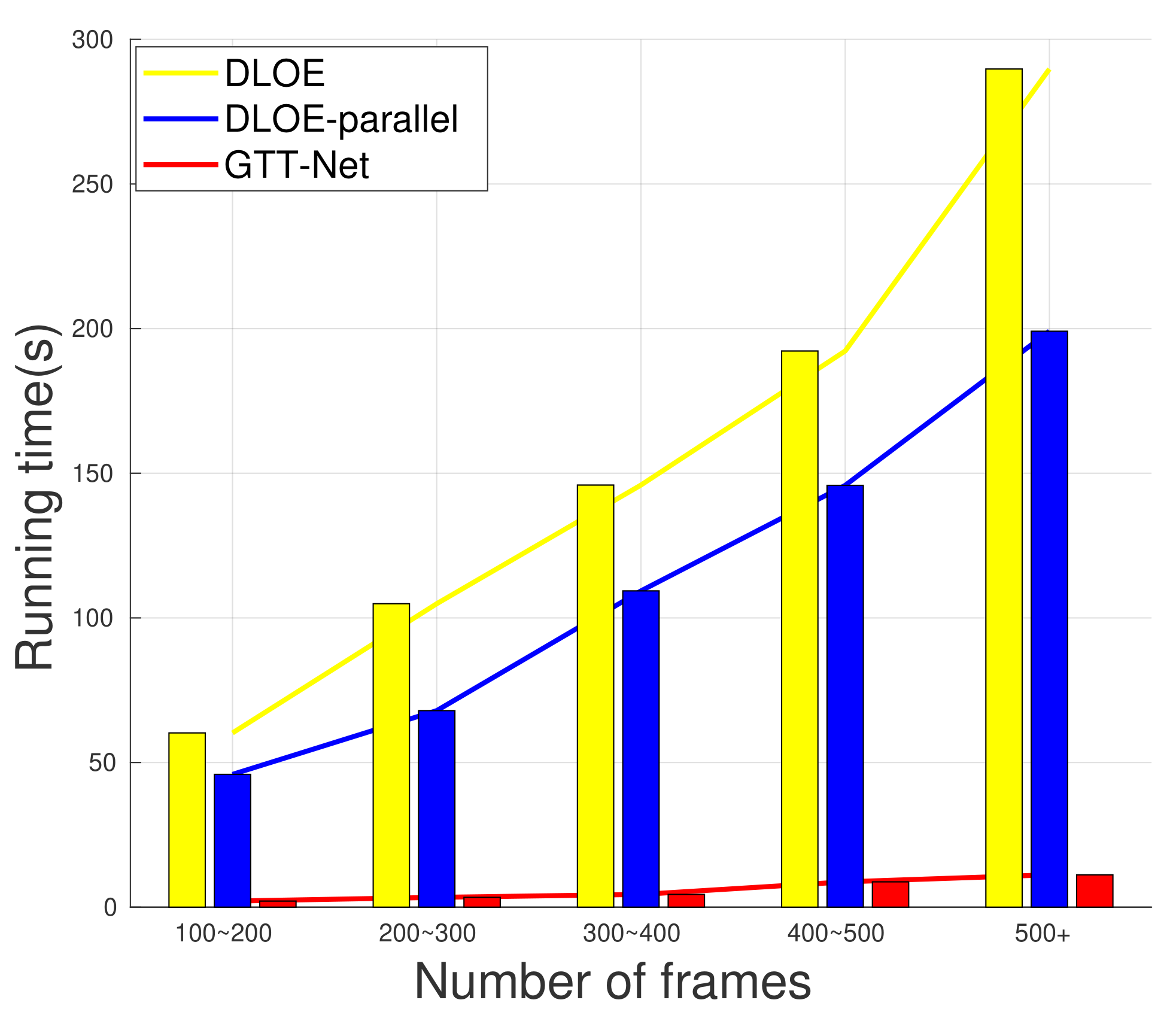}
        \caption{Efficiency comparison}
        \label{fig:timecompare}
    \end{subfigure}}
    \fbox{\begin{subfigure}[t!]{0.2\textwidth}
        \centering
        \includegraphics[scale=0.028]{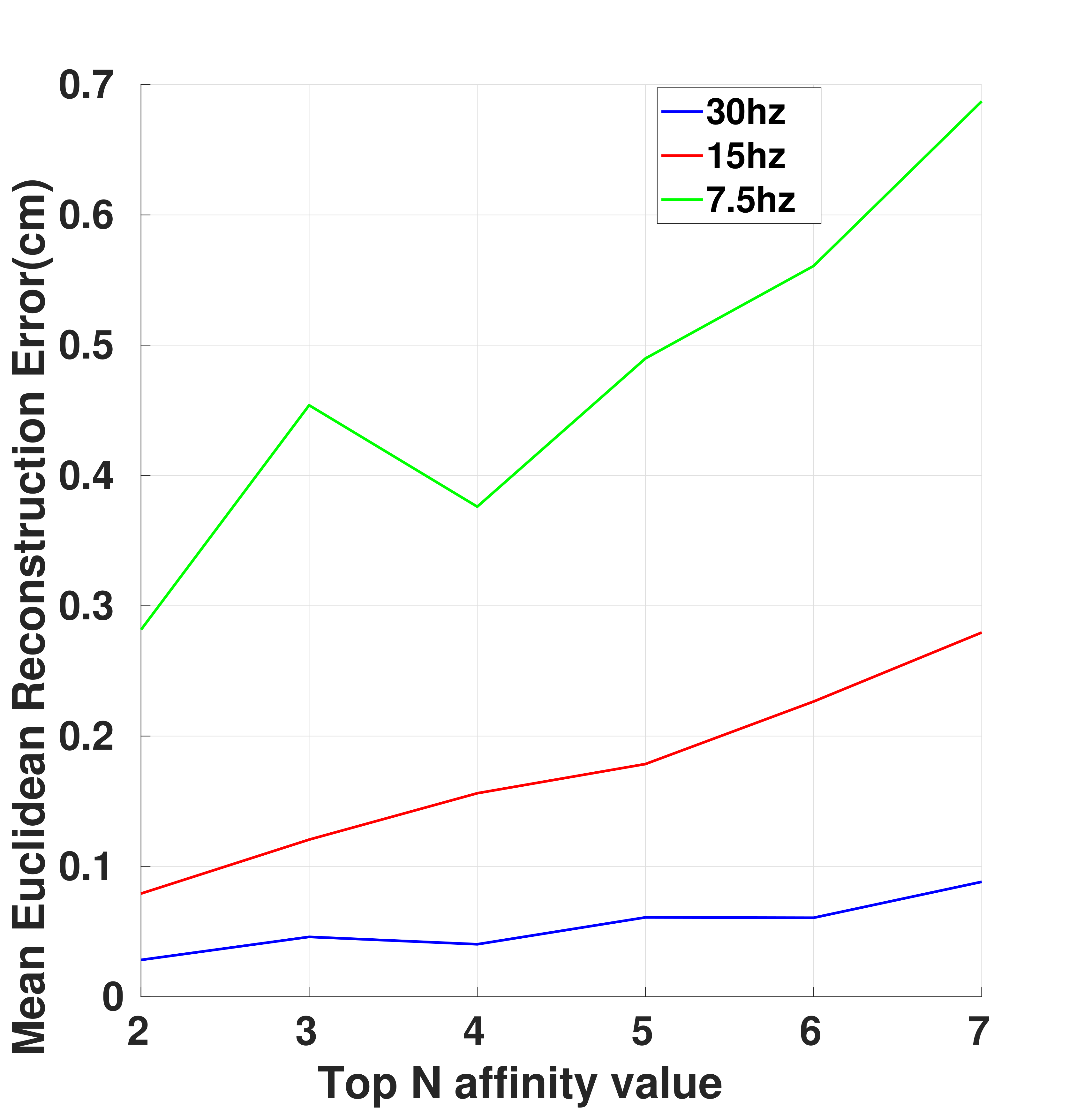}
        \caption{ Sparsity enforcement}
        \label{fig:N affinity}
    \end{subfigure}}
    \caption{(a) Computational efficiency comparision with DLOE \cite{Xu_2019_ICCV}. (b) Reconstruction error for different sparsity levels (i.e. keeping top N affinity value for each row).} 
\end{figure}
\subsection{Motion Capture Datasets}
%-------------------------------------------------------------------------
We use motion capture data \cite{cg-2007-2}  of 130 human 3D motions for 31 joints with frame rates of 120 Hz. We choose 10 of the 130 motions, each averaging  $\sim$300 frames for testing. We generate training datasets by randomly choosing from the remaining 120 datasets with varying levels of 2D noise, frames rates and percent of joints missing. We simulate four virtual cameras with $1000 \times 1000$ resolution and 1000 focal length. Dynamic 3D joint positions are projected  on them as 2D observations at a distance of 3m. For each 3D motion in both training and testing datasets, temporal sampling is performed at 30Hz and concurrent observations are systematically avoided to ensure all cameras are unsynchronized. We show results of 3D reconstruction accuracy comparisons in Fig. \ref{fig:Accuracy}.
GTT-Net is compared against discrete Laplace operator estimation (DLOE)\cite{Xu_2019_ICCV}, self-expressive dictionary learning (SEDL)\cite{zheng2017self}, trajectory basis (TB)\cite{park20153d}, high-pass filter (HPF)\cite{valmadre2012general} and the pseudo-triangulation approach in Sec.\ref{supervisory}. SEDL requires partial sequencing information. TB and HPF  require complete ground truth sequencing.

\noindent {\bf Varying 2D noise}. We randomly add 2D Guassian noise with standard deviation from 1 to 5 pixels to our observations. Fig. \ref{fig:DiffNoise} shows  GTT-Net is competitive with other methods across all sequencing information conditions. When full sequencing info is available, GTT-Net outperforms geometry-only methods (e.g. DLOE), indicating we learn improved affinity relations to triangulate 3D trajectories. Even without any sequencing info, GTT-Net  outperforms methods leveraging global sequencing. \\
%------------------------------------------------------------------------
\noindent {\bf  Varying frame rates}. We simulate lower frame rate conditions by downsampling the 2D capture to 7.5Hz, 15Hz and 30Hz. As shown in Fig. \ref{fig:DiffFreq}, our method performs better than DLOE on the conditions of partial sequencing information and no sequencing information. Working with full sequencing information, our method is still competitive.

%------------------------------------------------------------------------
\noindent {\bf Missing data}. We randomly decimate 3D joints at rates varying from 10\% to 50\%, and compare GTT-Net's  robustness against missing and/or occluded input features, see Fig. \ref{fig:DiffMissing}. Only DLOE, SEDL and TB are able to operate  having missing joints. The robustness of GTT-Net is competitive in all sequencing information conditions. 

%------------------------------------------------------------------------
\noindent {\bf Ablation of cascaded training}. 
Fig. \ref{fig:ablation} compares the reconstruction error distribution among the different stages in our cascaded training strategy.
%including the first stage and the final stage of weakly supervised and fully supervised learning version. 
We include a self-supervised version using only $\ell^{\text{\AE}}$ and $\ell^{\mathcal{A}}$ loss functions without external data. Surprisingly, self-supervised training is strongly competitive with full training cascade results, although subject to grater variability.

%Results show the effectiveness of our loss functions and cascaded training strategy.

%------------------------------------------------------------------------
\noindent {\bf PointNet network validation}. 
The PointNet-enabled variant of GTT-Net is trained on  different datasets of articulated 3D motion, such as monkeys\cite{bala2020openmonkeystudio}, hands\cite{yuan2017bighand2} and humans\footnote{CMU Mocap ( http://mocap.cs.cmu.edu/)}, see  Fig. \ref{fig:pointnet_train}, all having different joint topology compared to the testing data. In Fig. \ref{fig:PointNetValid}, we compare the reconstruction error distribution of three GTT-Net variants:
1) a Multi-domain PointNet-enabled GTT-Net, 2) a Single-Domain 3D-Supervised GTT-Net and 3) a Single-Domain 3D-Supervised GTT-Net where random rigid motions are applied to individual joint 3D trajectories to decorrelate their motion from the original motion semantics.
%a fully supervised version with additional version training on some dynamic 3D points without any semantic relation. 
Our PointNet variant outperforms the variant training on decorrelated input 3D motion and is competitive with the Single-Domain 3D-Supervised variant even though our PointNet variant was not exposed to the test domain during training.
%trained on 3D motions from  different domains.
The fact that training on decorrelated motion  provides inferior performance, indicates our GTT-Net framework effectively learns to enforce general 3D motion semantics when estimating inter-shape affinities.

%is better than only training on independent trajectories and close to fully supervised version which means that the PointNet nework does learn some general semantic information from the structure of different classes.

%------------------------------------------------------------------------
{ \noindent \bf Computational advantages over \cite{Xu_2019_ICCV}:}
 GTT-Net is over an order of magnitude  ($\sim30$X average speedup) faster than the open-source version of \cite{Xu_2019_ICCV} when estimating a single full-graph affinity matrix across different sequence lengths, while consistently being more accurate as in Fig. \ref{fig:timecompare}.

%------------------------------------------------------------------------
\begin{figure*}[t!]
    \fbox{\begin{subfigure}[t!]{0.485\textwidth}
        \centering
        \includegraphics[scale=0.245]{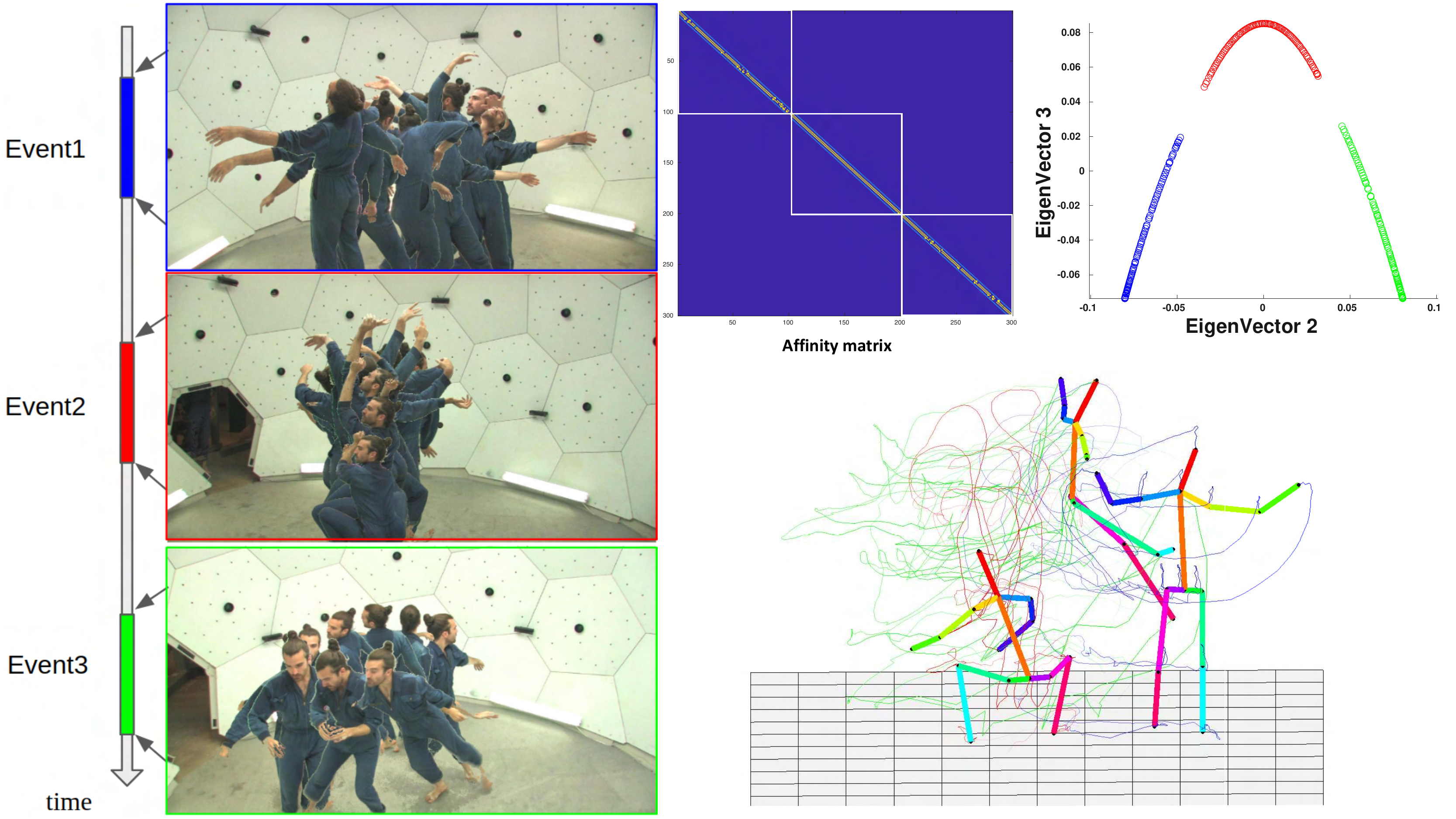}
        \caption{Multi-events segmentation}
        \label{fig:EventSegmentation}
    \end{subfigure}}
    \fbox{\begin{subfigure}[t!]{0.485\textwidth}
        \centering
        \includegraphics[scale=0.245]{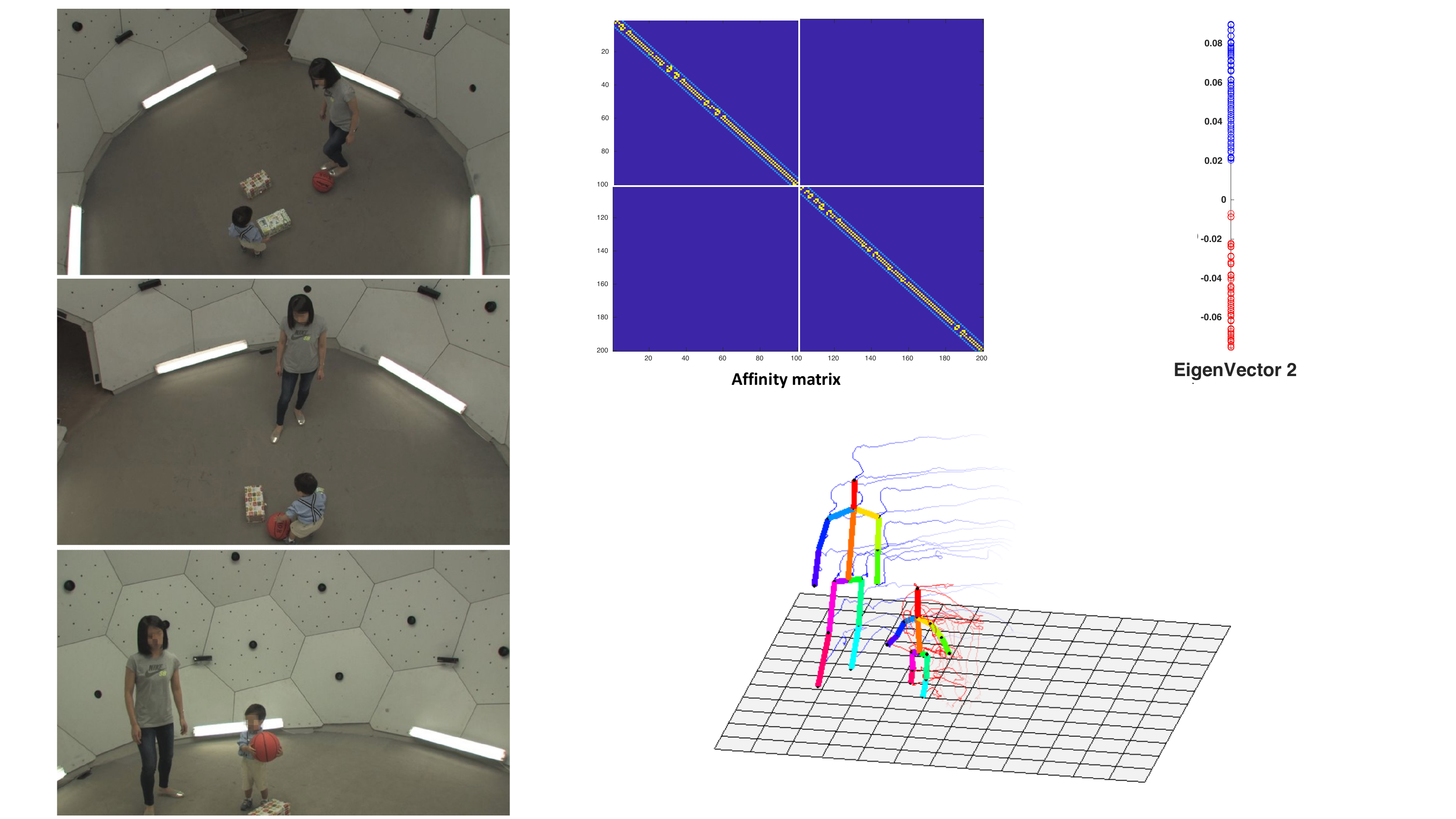}
        \caption{Multi-person}
        \label{fig:Multi-person}
    \end{subfigure}}
    \caption{(a) Result on a disjoint dancing scene. The affinity matrix and spectral analysis show the three segments  (b) This scenario includes an adult and a toddler which are clustered by the affinity matrix and corresponding spectral analysis. } 
\end{figure*}
\subsection{Cross-Domain Multi-view Video Datasets} 
\begin{figure*}[h!] 
    \fbox{\begin{subfigure}[h!] {0.485\textwidth} % width of left 
        \centering
		\includegraphics[scale = 0.25,trim={0cm 1.3cm 2cm 0cm},clip]{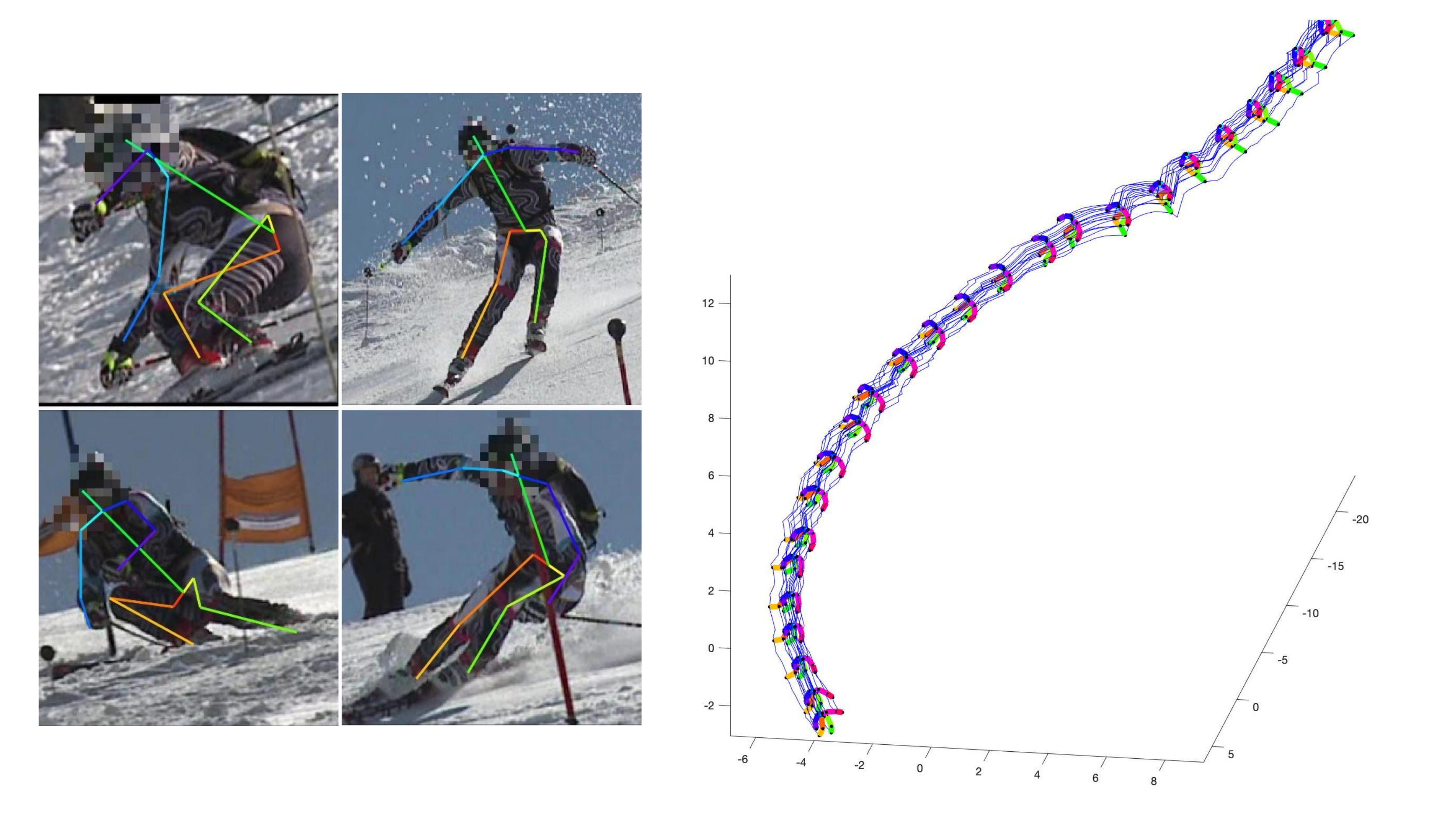}
		\label{fig:Ski}
		\caption{Ski}
	\end{subfigure}}
	\fbox{\begin{subfigure}[h!] {0.485\textwidth} % width of left 
        \centering
		\includegraphics[scale = 0.2335,trim={2cm 0cm 1.5cm 0cm},clip]{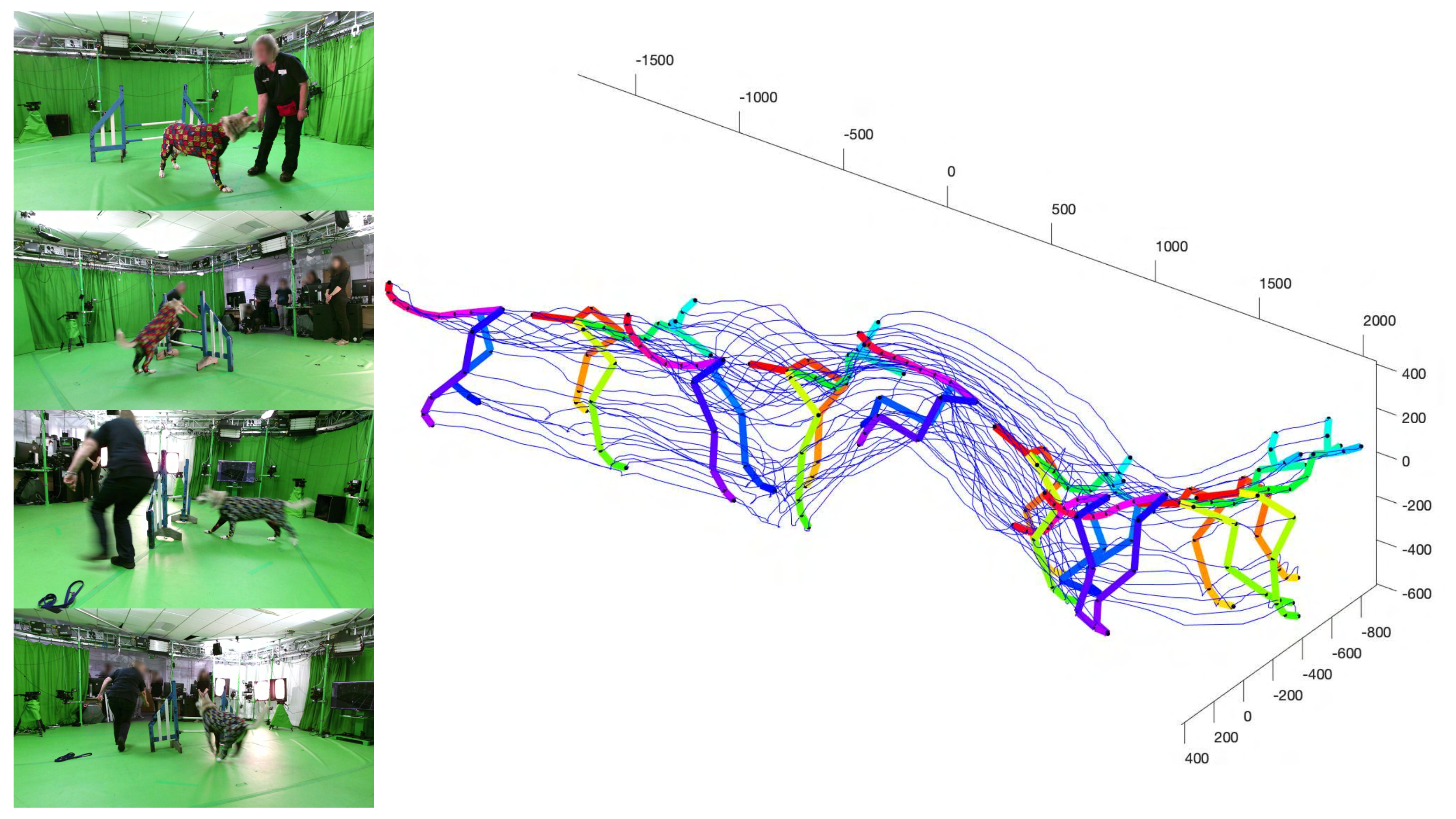}
		\label{fig:Dog}
		\caption{Dog}
	\end{subfigure}}
	\caption{Qualitative results on unsynchronized multi-view video capture. GTT-Net was not trained on the test domain. }
	\label{fig:real_exp}
\end{figure*}

Experiments on different   3D shapes classes illustrate the generality of our PointNet-enabled GTT-Net variant. The multi-view Human Ski \cite{rhodin2018learning} and Dog \cite{Kearney_2020_CVPR} datasets were unsynchronized and their provided 2D features were input to GTT-Net. Fig. \ref{fig:real_exp} illustrates our qualitative  results.
 GTT-Net was not exposed to either test domain during training.

%with known camera geometry and multi-view synchronized videos. For each dataset, we unsynchronized the multi-view video by selecting only one of the multiple images having a common timestamp. Since the structure and number of joints are different for each dataset, we utilize the trained PointNet variant of our network. Fig. \ref{fig:real_exp} illustrates our qualitative  results.

\subsection{Panoptic Studio dataset}  
CMU Panoptic Studio dataset \cite{joo2017panoptic} contains synchronized multi-view videos, 2D human joint estimates and camera poses. We sample video frames to generate multi-view unsynchronized image streams. Again, as the dataset-provided sparse shape feature inputs (i.e. skeleton joints)  are different from the 31-dimensional sparse shape features used for training,  we use the PointNet variant of GTT-Net.

\noindent {\bf Application to Event Segmentation}.
For multi-view videos capturing temporally separated events, our goal is to jointly reconstruct the dynamic 3D structure and segment all events based on the estimated affinity matrix $\mathbb{A}^\mathcal{S}$. $\mathbb{A}^\mathcal{S}$  described a graph with multiple connected components, each of which corresponds to a separate event.
For each segmented event, the sequencing of its constituting images  was directly extracted from the affinity matrix. From top right in Fig. \ref{fig:EventSegmentation}, we can notice the chain-like structure for each event by performing spectral analysis on the affinity matrix.

\noindent {\bf Application to Multi-Target Scenarios}.
We consider the case where multiple shapes are captured in multi-view cameras, but the shape's correspondence among the images is not available. Given $N$ images $\{{\mathcal{I}}_{n}\}$  with maximal $M$ shape captured, the goal is to reconstruct the aggregated dynamic 3D structure $\mathbb X_{i,:} \in \mathbb R^{3MP}$.% and $M$ objects clustering. 
We propose a solution for this case based on GTT-Net: {\bf 1}) We separately create virtual frames \{$\tilde{\mathcal{I}}_{n,m}\}$(each observing $P$ 3D points) for each of the subjects captured in original images. 
{\bf 2}) Execute GTT-Net on the (up to $NM$) new virtual images to reconstruct the 3D structure and generate the corresponding affinity matrix as in the single shape case. {\bf 3}) Cluster individual objects based on the affinity matrix by any standard  clustering method. {\bf 4}) Merge estimated 3D shapes originating from the same image.  {\bf 5}) Run GTT-Net on the $N$ original input images with aggregated shape feature  to refine the decoupled event reconstructions from step 2.  
Fig. \ref{fig:Multi-person} shows our  results for a two-target scenario.

%----------------------------------------------------------------------------
\section{Conclusion}
GTT-Net uses supervised learning to  estimate  pairwise spatio-temporal affinities and compute  dynamic 3D  geometry from image observations. Our framework allows for a diverse set of training scenarios and leverages them on a cascaded supervision strategy to both improve training efficiency and be adaptive to available supervisory information. Moreover, the proposed system is robustly applicable across different shape-trajectory domains, while outperforming the current state of the art.  
%------------------------------------------------------------------------

{\small
\bibliographystyle{ieee_fullname}
\bibliography{egbib}
}

\end{document}